\documentclass{article} % For LaTeX2e
\usepackage{iclr2026_conference,times}

\usepackage{float}

\usepackage{amsmath,amsfonts,bm}

\def\eqref#1{equation~\ref{#1}}
\def\1{\bm{1}}

\DeclareMathAlphabet{\mathsfit}{\encodingdefault}{\sfdefault}{m}{sl}
\SetMathAlphabet{\mathsfit}{bold}{\encodingdefault}{\sfdefault}{bx}{n}

\usepackage{amsmath, amsthm, amssymb}
\newtheorem{theorem}{Theorem}
\newtheorem{lemma}{Lemma}[section]
\theoremstyle{definition}
\newtheorem{definition}{Definition}
\usepackage{xcolor}

\usepackage{booktabs}
\usepackage{multirow}
\usepackage{graphicx}
\usepackage[table]{xcolor}
\definecolor{rowgray}{gray}{0.96}
\usepackage{wrapfig}
\usepackage{subcaption}

\usepackage{enumitem}

\usepackage[ruled,vlined,linesnumbered]{algorithm2e}

\usepackage[utf8]{inputenc}
\usepackage[T1]{fontenc}
\usepackage{amsfonts}
\usepackage{nicefrac}
\usepackage{microtype}
\usepackage{comment}
\usepackage{url}

\usepackage{hyperref}

\title{PiCa: Parameter-Efficient Fine-Tuning\\with Column Space Projection}

\iclrfinalcopy

\author{
  Junseo Hwang\thanks{Equal contribution.} \quad
  Wonguk Cho\footnotemark[1] \quad
  Taesup Kim\thanks{Corresponding author.} \\
  Graduate School of Data Science, Seoul National University \\
}

\begin{document}

\maketitle
\begin{abstract}
Fine-tuning large foundation models is essential for building expert models tailored to specialized tasks and domains, but fully updating billions of parameters is computationally prohibitive. Reducing the number of trainable parameters using Parameter-Efficient Fine-Tuning (PEFT), such as Low-Rank Adaptation (LoRA), is therefore crucial not only to reduce training costs but also to mitigate storage, caching, and serving overheads during deployment. Prior works, such as Singular Vectors-guided Fine-Tuning (SVFT), have shown that exploiting the geometry of pre-trained weights based on Singular Value Decomposition (SVD) can significantly improve parameter-efficiency, but they lack a solid theoretical foundation. In this paper, we introduce Parameter-Efficient Fine-Tuning with Column Space Projection (PiCa), a novel theoretically grounded PEFT method. We prove that projecting gradients onto the principal column space of pre-trained weights provides an effective inductive bias for adaptation and further enhance parameter efficiency through a novel weight-sharing strategy. Across diverse NLP and vision tasks, PiCa consistently outperforms state-of-the-art baselines under comparable or smaller parameter budgets, demonstrating both theoretical rigor and practical effectiveness. 
\end{abstract}

\section{Introduction}

Fine-tuning large foundation models is essential for building expert models tailored to specialized tasks and domains. However, fully fine-tuning billions of parameters is often computationally prohibitive in terms of both training and deployment cost. Parameter-Efficient Fine-Tuning (PEFT)~\citep{houlsby2019parameter} addresses this challenge by adapting models with only a small number of trainable parameters while keeping the pre-trained backbone frozen. In particular, minimizing the number of trainable parameters is critical in practical scenarios where multiple adapters must be deployed simultaneously~\citep{chen2024punica}. In such cases, numerous sets of fine-tuned parameters for different tasks, models, and checkpoints per user must be stored separately from the pre-trained models, leading to significant storage, caching, and serving overheads.

A prominent line of research is Low-Rank Adaptation (LoRA)~\citep{hu2022lora}, known for its simplicity and strong empirical performance. While reducing its rank lowers the number of trainable parameters, it inevitably causes significant performance degradation. To address this, DoRA~\citep{liu2024dora} introduces weight decomposition into LoRA, achieving stronger performance at a fixed rank and often matching or surpassing LoRA while requiring only half the trainable parameters. VeRA~\citep{kopiczko2023vera} further reduces parameter budgets by training small scaling vectors, demonstrating that comparable or superior performance to LoRA can be obtained with up to 4$\times$ fewer trainable parameters. However, these LoRA-based methods typically rely on randomly initialized low-rank matrices and thus do not explicitly leverage the geometry or prior knowledge encoded in the pre-trained weights.

Furthermore, recent studies~\citep{lingam2024svft,han2023svdiff,mantri2025ditask} have shown that leveraging the geometry of pre-trained weights, particularly their spectral structure, can lead to further parameter-efficiency without performance degradation. For instance, Singular Vectors-guided Fine-Tuning (SVFT)~\citep{lingam2024svft} constructs a sparse, weighted combination of a model’s pre-trained singular vectors to achieve strong performance with fewer trainable parameters. However, despite their empirical success, these SVD-based approaches~\citep{lingam2024svft,han2023svdiff,mantri2025ditask} lack theoretical foundation for their approaches and leave open why using the spectral structure of pre-trained weights constitutes an effective inductive bias for fine-tuning.

\begin{wrapfigure}{r}{0.56\textwidth} \centering \includegraphics[width=0.56\textwidth]{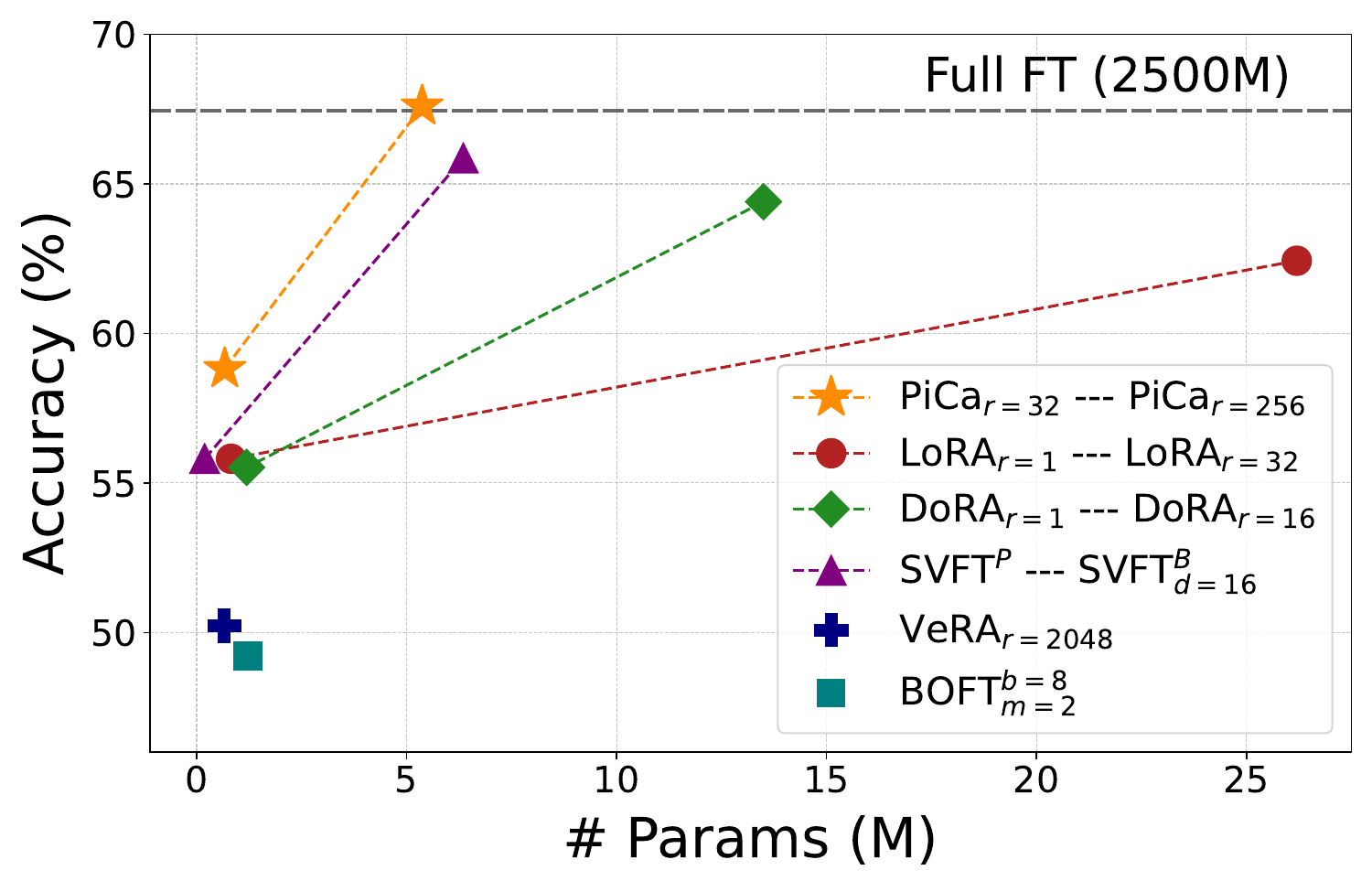} 
\vspace{-18pt}
\caption{Average accuracy as a function of the number of trainable parameters on Commonsense Reasoning datasets using Gemma-2B. PiCa demonstrates superior performance compared to baseline methods with similar parameter budgets.} 
\label{fig:fig2} 
\vspace{-12pt}
\end{wrapfigure} 

In this work, we propose \textbf{P}arameter-Eff\textbf{i}cient Fine-Tuning with \textbf{C}olumn Sp\textbf{a}ce Projection (\textbf{PiCa}), \textit{a new theoretically grounded PEFT method that leverages the geometry of pre-trained weights.} Our theoretical analysis demonstrates that projecting gradients onto the principal column space spanned by pre-trained weights can lead to effective adaptation. This gradient projection is effectively paired with our novel weight-sharing method for further parameter efficiency. With this approach, we can significantly reduce the number of trainable parameters, even using less than the most parameter-efficient configurations of other methods (e.g., rank-1 LoRA and DoRA), while achieving significantly better performance. Our extensive experiments across various models and datasets demonstrate that PiCa consistently outperforms all baseline methods under comparable parameter budgets, as shown in Fig~\ref{fig:fig2}.

Our contributions can be summarized as follows:
\begin{itemize}[leftmargin=2em]
\item We introduce \textbf{PiCa},  a \textbf{theoretically grounded PEFT method} that explicitly exploits the geometry of pre-trained weights. We provide \textbf{a theoretical foundation} showing that projecting gradients onto the principal column space of pre-trained weights enables effective adaptation. For further parameter efficiency, PiCa also introduces a \textbf{novel weight-sharing approach} that can be paired with gradient projection.

\item \textbf{PiCa consistently achieves competitive or superior performance with significantly fewer parameters} compared to other baselines. In particular, it \textbf{outperforms state-of-the-art baselines}, SVFT$^{R}$ and SVFT$^{B}$, across all datasets and models under smaller parameter budgets.

\item Our experiments span a \textbf{wide range of NLP tasks} including mathematical reasoning, commonsense reasoning, and natural language understanding with different language models, as well as \textbf{diverse vision tasks} such as visual adaptation on 19 VTAB datasets with vision transformers and subject-driven generation on DreamBooth with text-to-image diffusion models. We also conduct \textbf{comprehensive ablation studies} to better understand the individual components of our method and their effects.
\end{itemize}

\section{Related Work}
\paragraph{Parameter-Efficient Fine-Tuning}
In adapting large foundation models for downstream tasks, while full fine-tuning often yields superior performance on these tasks, its prohibitive computational overheads have motivated the development of various PEFT methods that aim to achieve comparable performance with much smaller number of trainable parameters. Recently highlighted approaches include low rank approximation~\citep{hu2022lora, liu2024dora, kopiczko2023vera}, orthogonal reparametrization~\citep{qiu2023controlling,boft}, and Singular Value Decomposition (SVD)-based approaches~\citep{lingam2024svft, han2023svdiff,mantri2025ditask}. 

In particular, LoRA and its variants~\citep{hu2022lora, liu2024dora, kopiczko2023vera} have significant attention due to its simplicity and efficiency, based low-rank decomposition. DoRA~\citep{liu2024dora} decomposes weights and achieves stronger performance at a fixed rank, often matching or surpassing LoRA while requiring only half the trainable parameters. VeRA~\citep{kopiczko2023vera} further reduces parameter budgets by training small scaling vectors.

On the other hand, methods leveraging the structure of pre-trained weights, specifically through their SVD components, have been explored~\citep{lingam2024svft, han2023svdiff,mantri2025ditask,meng2024pissa}.  SVFT~\citep{lingam2024svft} utilizes the entire singular vectors of pre-trained weights as a basis and employs a sparse matrix for updates. SVDiff~\citep{han2023svdiff} has demonstrated fine-tuning only the singular values of pre-trained weight matrices is effective in personalization of text-to-image diffusion models. Similarly, DiTASK~\citep{mantri2025ditask} has shown that preserving singular vectors and enabling task-specific adaptations through neural diffeomorphic transformations of the singular values can be effective for dense prediction tasks.

Although these SVD-based methods have shown empirical success, they often lack a strong theoretical foundation that provides an analytical justification for their methods, and only few works has attempted to analyze the change in spectral structure after fine-tuning~\citep{shuttleworth2024lora}. In contrast, we develop a method based on a theoretical proof that the optimal rank-$r$ approximation of $\Delta W$ can be achieved by the singular vectors of the pre-trained weights, which aligns with our empirical findings. The effectiveness of this approach is validated through extensive experiments.

\paragraph{Weight-sharing}

Prior research has explored weight-sharing to reduce the number of parameters in neural networks~\citep{press-wolf-2017-using,inan2016tying}. More recently, this concept of weight-sharing has been adapted within the LoRA framework~\citep{kopiczko2023vera,renduchintala2023tied, zhou2025bishare,shen2024sharelora,song2024sharelora}. For instance, VeRA~\citep{kopiczko2023vera} introduces a frozen random projection matrix shared across all layers, combined with trainable scaling vectors. Furthermore, recent works~\citep{renduchintala2023tied,song2024sharelora} explore different strategies of combining freezing, training, and sharing both projection matrices and scaling vectors. While demonstrating progress in parameter reduction, these prior approaches tend to be highly sensitive to randomly initialized projection matrices and often their performance is below that of standard LoRA. However, in PiCa, we construct projection matrix based on structure of pre-trained weights for each layer and share trainable weights across layers with the same function role. This approach allows significant reduction of trainable parameters without performance degradation.

\section{Methodology}
In this section, we introduce our novel PEFT method, PiCa. (1) We first discuss how fine-tuning relates to singular vectors and introduces Theorem~\ref{thm:pica}, which shows that the principal subspace of pre-trained weights offers an effective space for adaptation (Section~\ref{3.1}). (2) We develop this idea in the context of PEFT settings, showing that sequentially projecting gradients onto this subspace offers a theoretically grounded way to perform fine-tuning under parameter constraints through Theorem~\ref{thm:sequential} (Section~\ref{3.2}). (3) On top of these insights, we finally present our algorithm, PiCa, which integrates sequential projection with weight-sharing for further parameter-efficient adaptation (Section~\ref{3.3}).

\subsection{Fine-tuning and Column space projection}
\label{3.1}
\noindent
\begin{minipage}[t]{0.45 \columnwidth}
“Fine"-tuning is, by definition, the process of making a relatively small update
from the pre-trained weights $W_0$ to the fine-tuned weights $W^*$, in order to adapt
a model to a specific downstream task with a much smaller dataset. As large foundation
models are pre-trained on vast, diverse corpora, good optima tend to lie in a small-update
neighborhood of $W_0$. Therefore, in the context of fine-tuning of large foundation models,
it is natural to assume that $\Delta W = W^* - W_0$ with $\|W_0\| \gg \|\Delta W\|$.
Lemma~\ref{lemma:wedin} indicates that, when this change is small, the leading singular
structures of $W_0$ and $W^*$ remain closely aligned.
\end{minipage}\hfill
\begin{minipage}[t]{0.52\columnwidth}
\vspace{-8pt}
\centering
\includegraphics[width=\linewidth]{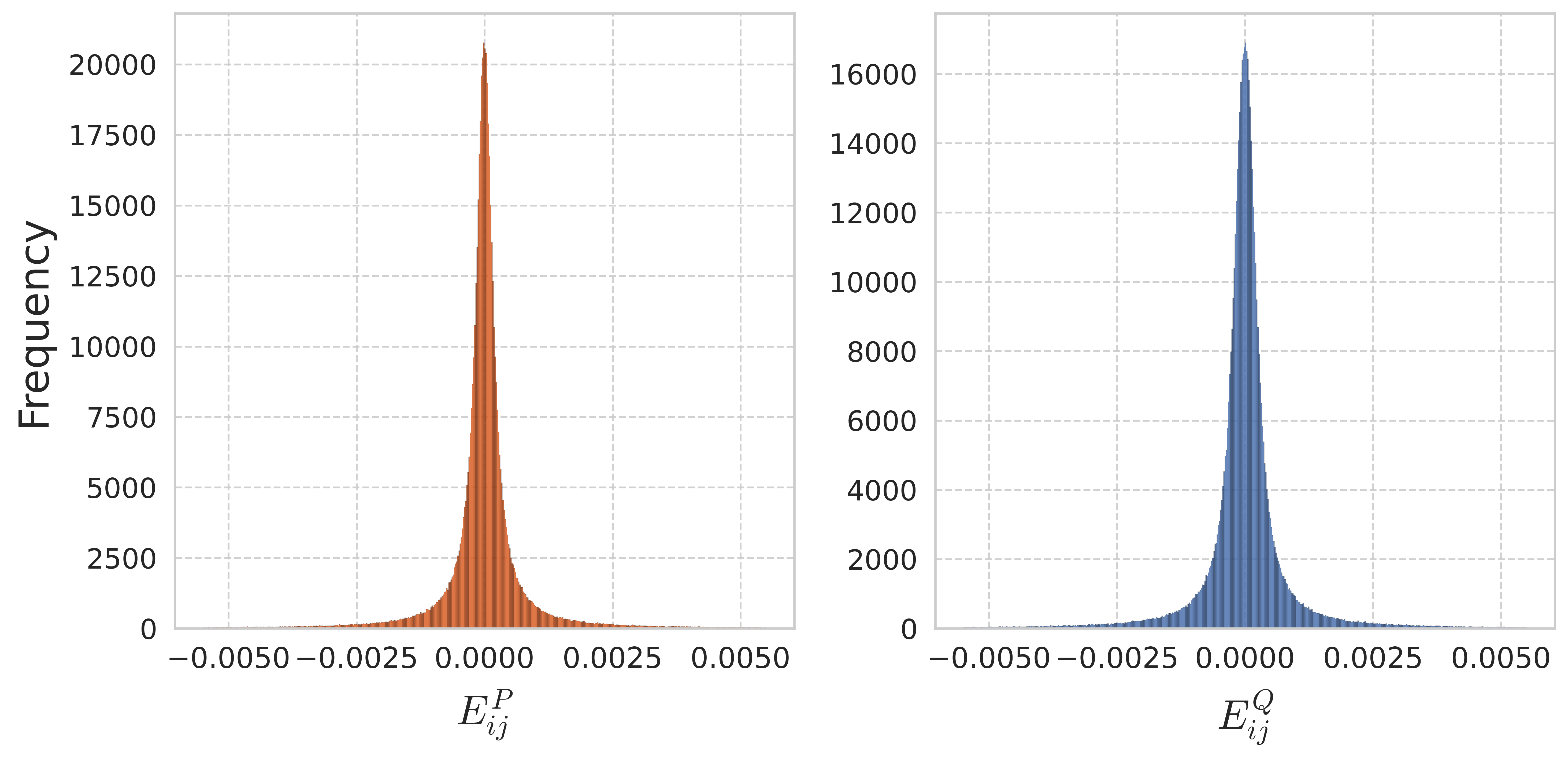}
\captionof{figure}{Distribution of \(E_{ij}^P\) and \(E_{ij}^Q\) across all weight matrix elements using DeBERTaV3\textsubscript{base}. Most values are tightly concentrated around zero, validating that \(\mathcal{O}(\epsilon)\) is negligible in practice.}
\label{fig:fig3}
\end{minipage}

\begin{lemma}[\cite{wedin1972perturbation}]
\label{lemma:wedin}
Let $W_0,W^*\in\mathbb{R}^{m\times n}$ with $W^*=W_0+\Delta W$. 
Let $U_r,U_r^*$ denote the top-$r$ left singular-vector matrices of $W_0$ and $W^*$. 
Define the gap
\[
\delta := \min\!\Big\{\sigma_r(W_0)-\sigma_{r+1}(W^*),\ \sigma_r(W^*)-\sigma_{r+1}(W_0)\Big\}.
\]
Then for any unitarily invariant norm,
\[
\|\sin\Theta(U_r,U_r^*)\| \;\le\; \frac{\|\Delta W\|}{\delta}.
\]
\end{lemma}

Building on this insight, Theorem~\ref{thm:pica}  expresses the relation between $W_0$ and $W^*$ in a form that involves a small deviation $E$, and uses this to analyze how the update $\Delta W$ can be captured within the column space of $U_r$. Empirical results in Fig.~\ref{fig:fig3} support this view, showing that the entries of $E$ are tightly concentrated near zero.

\vspace{0.5cm}
\begin{theorem}[Approximation error of projection onto $U_r$]
\label{thm:pica}
Let \( W_0 = U \Sigma V^\top \in \mathbb{R}^{m \times n} \) be the Singular Value Decomposition (SVD) of \( W_0 \).  
Suppose the fine-tuned matrix \( W^* \in \mathbb{R}^{m \times n} \) has the form
\[
W^* = (U P) \Sigma^* (V Q)^\top,
\]
where:
\begin{itemize}
    \item \(U^*=UP \) and \(V^* =VQ \) are the left and right singular vectors of \( W^* \), respectively,
    \item \( \Sigma^* = \text{diag}(\sigma_1(W^*), \dots, \sigma_{\min(m,n)}(W^*)) \),
    \item \( P = I_m + E^P \), \( Q = I_n + E^Q \), with \( |E^P_{ij}| < \epsilon \), \( |E^Q_{ij}| < \epsilon \).

\end{itemize}
Let \( \Delta W = W^* - W_0 \), and let \( U_r \in \mathbb{R}^{m \times r} \) be the top-\( r \) left singular vectors of \( W_0 \).  
Then, the approximation error incurred by projecting \( \Delta W \) onto the subspace spanned by \( U_r \) satisfies
\[
\left\| \Delta W - U_r U_r^\top \Delta W \right\|_F^2 \leq \sum_{i=r+1}^{\min(m,n)} \sigma_i^2(\Delta W) + \mathcal{O}(\epsilon).
\]
\end{theorem}
\noindent The complete proof of Theorem~\ref{thm:pica} is detailed in Appendix~\ref{sec:proof}.

Theorem~\ref{thm:pica} indicates that the update $\Delta W$ can be well captured within the principal column space of $W_0$. The first term on the right-hand side, $\sum_{i=r+1}^{\min(m,n)}\sigma_i^2(\Delta W)$, corresponds to the rank-$r$ approximation error of $\Delta W$ given by the Eckart–Young theorem~\citep{eckart1936approximation}. The additional $\mathcal{O}(\epsilon)$ term reflects the small deviation introduced through $E^P$ and $E^Q$, and empirical evidence in Fig.~\ref{fig:fig3} suggests that 
the \( \mathcal{O}(\epsilon) \) term is negligible in practice. Appendix~\ref{app:evidence_llama} provides further observations on large-scale models, which is consistent with this view.

Theorem~\ref{thm:pica} shows that the dominant directions of the resulting update $\Delta W$ are well captured within the pre-trained column space $U_r$ of $W_0$. This implies that by keeping $U_r$ fixed and learning only a small set of coefficients that determine the task-specific choice of how to move inside this space, we can substantially reduce the number of trainable parameters, which is precisely the notion of parameter efficiency we target.

Theorem~\ref{thm:pica} is not meant to show that $U_r$ projection is globally optimal or that projection alone guarantees task-optimal performance. Other projection spaces may also reach good optima, which does not contradict our claims. Rather, our contribution is to provide theoretical support for why this particular projection can work well, whereas most prior methods are justified only empirically.

\subsection{Sequential Gradient Projection}
\label{3.2}
Theorem~\ref{thm:sequential} shows that the principal column space in Theorem~\ref{thm:pica} can be naturally incorporated into PEFT by projecting gradients onto the subspace at each step. This provides a practical way to exploit the same effective space throughout training, offering a simple and theoretically supported view of how sequential updates can operate within the projection framework.

\begin{definition}[L-smoothness for matrix-valued functions]
\label{def:matrix-Lsmooth}
A differentiable function $\ell:\mathbb{R}^{m\times n}\to\mathbb{R}$ is \emph{$L$-smooth (w.r.t.\ $\|\cdot\|_F$)} if
\[
\|\nabla \ell(W_1)-\nabla \ell(W_2)\|_F \;\le\; L\,\|W_1-W_2\|_F
\quad\text{for all }W_1,W_2\in\mathbb{R}^{m\times n}.
\]
\end{definition}

\begin{theorem}[Sequential projection approximates accumulated projection]
\label{thm:sequential}
Let $\ell:\mathbb{R}^{m\times n}\to\mathbb{R}$ be $L$-smooth with $\|\nabla \ell(W)\|_F\le G$.  
Define the unprojected gradient descent path
\[
Z_{t+1}=Z_t-\eta\nabla \ell(Z_t).
\]
Let the \emph{accumulated-projection} iterate be
\[
W_T
= W_0 - \eta\,\Pi_{U_r}\!\Bigl(\sum_{t=0}^{T-1}\nabla \ell(Z_t)\Bigr),
\]
and the \emph{sequential-projection} iterates
\[
P_{t+1}=P_t-\eta\,\Pi_{U_r}\nabla \ell(P_t), 
\qquad P_0=W_0,
\]
where $\Pi_{U_r}=U_r U_r^\top$ is the fixed rank-$r$ projector.  

Then, for any $T$, the difference satisfies
\[
\|W_T-P_T\|_F
\;\le\;\frac{\eta^2}{2}\,L G\,T(T-1)
+ O((\eta L T)^3).
\]
\end{theorem}
\noindent The proof is provided in Appendix~\ref{sec:proof}.

\begin{algorithm}[t]
\LinesNotNumbered
\caption{Adam with PiCa}
\label{alg:pica}
\KwIn{rank $r$; learning rate $\eta$; decay rates $\beta_1,\beta_2$; small $\varepsilon>0$.}
\BlankLine
\textbf{Setup / Notation.} For each group $f \in \mathcal{F}$ and layer $i$: compute SVD 
$W^{f,i}_0 = U^{f,i}\Sigma^{f,i}(V^{f,i})^\top$ and set $P^{f,i} \gets U^{f,i}_{[:,1:r]}$ \tcp*{Layer-wise fixed projector}
Set $W^{f,i} \gets W^{f,i}_0 \in \mathbb{R}^{m \times n}$ \;
For each group $f$: set shared compact states 
$B^f_0,M^f_0,V^f_0 \in \mathbb{R}^{r\times n}\gets 0$; set $t\gets 0$\;
\textit{Elementwise ops:} $\odot$ (Hadamard), $\oslash$ (elementwise divide), $\sqrt{\cdot}$ (elementwise).
\BlankLine
\Repeat{convergence}{
  $t\gets t+1$ \;
  \ForEach{group $f$}{
    \tcp{(1) Project layer-wise gradients \& aggregate}
    $R^f_t \gets \sum_{i}\, \colorbox{gray!15}{$(P^{f,i})^\top$}\!\Big(-\nabla_{W^{f,i}}\,\ell_t(W^{f,i})\Big)$ \;
    \BlankLine
    \tcp{(2) Adam update in compact space}
    $M^f_t \gets \beta_1 M^f_{t-1} + (1-\beta_1)R^f_t$ \;
    $V^f_t \gets \beta_2 V^f_{t-1} + (1-\beta_2)(R^f_t\odot R^f_t)$ \;
    $\hat M^f_t \gets M^f_t/(1-\beta_1^{\,t})$; \quad
    $\hat V^f_t \gets V^f_t/(1-\beta_2^{\,t})$ \;
    $\Delta B^f_t \gets \hat M^f_t \oslash(\sqrt{\hat V^f_t}+\varepsilon)$; \quad
    $B^f_t \gets B^f_{t-1} + \eta\,\Delta B^f_t$ \;
    \BlankLine
    \tcp{(3) Decompress shared update to each layer}
    \ForEach{layer $i$}{
      $W^{f,i} \gets W^{f,i} + {\eta}\colorbox{gray!15}{$P^{f,i}$}\,\Delta B^f_t$ \;
    }
  }
}
\KwRet{$\{B^f_T\}_{f \in\mathcal{F}}$ \tcp*{Final shared compact parameters}}
\end{algorithm}

\vspace{0.5cm}
\subsection{PiCa: PEFT with Column Space Projection}
\label{3.3}
Based on the preceding results, we propose PiCa that projects gradients onto the principal column space spanned by pre-trained weights for each update. This gradient projection is effectively paired with our novel weight-sharing method for further parameter efficiency. For clarity, we describe PiCa in Algorithm~\ref{alg:pica} using Adam, though the approach is not limited to this optimizer.

In Algorithm~\ref{alg:pica}, each functional group $f \in \mathcal{F}=\{\text{query,key,value,\dots}\}$ is associated with a single trainable matrix $B^f \in \mathbb{R}^{r\times n}$, which is shared across all layers $i=1,\dots,L$ of the same group. The projection matrices $P^{f,i}$ remain layer-specific, leveraging the geometry of each pre-trained weight $W_0^{f,i}\in \mathbb{R}^{m \times n}$. The gradients of each layer $i$ are first projected onto $P^{f,i}$ defined by the top-$r$ singular vectors of the corresponding pre-trained weight, $U_r^{f,i}\in \mathbb{R}^{m \times r}$. The updates are then accumulated in this compact space as shared parameters $B^f$. Momentum and variance statistics are also updated in this compact space.
The shared update is then mapped back to each layer through its layer-specific projector $U_r^{f,i}$.

This procedure can be implemented using the reparameterization
\[
W^{f,i} = W_0^{f,i} + U_r^{f,i} B^f,
\]
where \( B^f \) is a trainable matrix initialized with zero and \(U_r^{f,i} \) remains fixed during fine-tuning.
% where $U_r^{f,i}$ is fixed and only $B^f$ is optimized, initialized to zero.
Under this parameterization, optimizing $B^f$ is mathematically equivalent to the update in Algorithm~\ref{alg:pica}.

Unlike prior approaches~\citep{kopiczko2023vera,renduchintala2023tied} that primarily rely on random projection matrices for weight-sharing,
our method leverages layer-specific projection matrices $U_r^{f,i}$ derived from the structure of the pre-trained weights $W_0^{f,i}$ for each layer $i$ of group $f$. This allows us to capture the distinct characteristics and pre-trained knowledge encoded in each $W_0^{f,i}$. Given the use of unique projection matrices per
layer, we posit that the trainable parameter $B^f$
can be effectively shared across layers with the same
functionality, facilitating efficient adaptation to downstream tasks. Our extensive experiments demonstrate the effectiveness of weight-sharing in PiCa, which reduces the number of trainable parameters by up to 7$\times$ without compromising performance (see Sec.~\ref{further} for details).

\section{Experiments}
\label{sec:exp}
\subsection{Experimental settings}

We evaluate the effectiveness of PiCa across a diverse set of Natural Language Processing (NLP) tasks, covering Mathematical Reasoning, Commonsense Reasoning, and Natural Language Understanding (NLU). For Mathematical Reasoning tasks, we fine-tune our model on the MetaMathQA-40K dataset~\citep{yu2023metamath} and assess its performance on the GSM-8K~\citep{gsm8k} and MATH~\citep{hendrycks2021measuring} datasets. Furthermore, we conduct evaluations on eight commonsense reasoning benchmarks: BoolQ~\citep{boolq}, PIQA~\citep{piqa}, SIQA~\citep{socialiqa}, HellaSwag~\citep{hellaswag}, Winogrande~\citep{winogrande}, ARC-Easy/ARC-Challenge~\citep{arc}, and OpenBookQA~\citep{openbookqa}. For NLU tasks, we utilize the GLUE benchmark~\citep{glue}. We report matched accuracy for MNLI, Matthew’s correlation for CoLA, Pearson correlation for STS-B, and accuracy for all other tasks. We employ the Gemma-2B/7B~\citep{gemma}, and LLaMA-3-8B~\citep{llama3} models for Mathematical Reasoning tasks and adopt the DeBERTaV3-base~\citep{debertav3} model for NLU tasks. 

Beyond NLP, we also evaluate PiCa on vision tasks. Specifically, we conduct experiments with visual adaptation using the ViT-B/16~\citep{dosovitskiy2021imageworth16x16words} on 19 different datasets of VTAB-1K~\citep{zhai2020largescalestudyrepresentationlearning}, grouped into \emph{Natural}, \emph{Specialized}, and \emph{Structured} categories. Performance is reported as the average accuracy across these groups. In addition, we evaluate subject-driven generation tasks with the Stable Diffusion v2.1~\citep{rombach2022highresolutionimagesynthesislatent} on the DreamBooth dataset~\citep{ruiz2023dreamboothfinetuningtexttoimage}, which includes 30 subjects and 25 prompts per subject, totaling 750 different personalization tasks. Following prior work~\citep{ruiz2023dreamboothfinetuningtexttoimage}, we report results using DINO for subject fidelity and CLIP-T for text fidelity.
To ensure a fair comparison, hyperparameters and training protocols are aligned with those outlined in ~\citep{lingam2024svft, cho2024hollowed}. Further details are provided in the Appendix~\ref{appendix:hyperparam}. %
\footnote{The official implementation is available at ~\url{https://github.com/hjunseoh/PiCa}.}

\subsection{Results} 
For a fair comparison, we follow \citep{lingam2024svft, dosovitskiy2021imageworth16x16words,cho2024hollowed} and evaluate the effectiveness of PiCa across three NLP tasks (Mathematical Reasoning, Commonsense Reasoning, and Natural Language Understanding) and two vision tasks (Visual Adaptation and Subject-Driven Generation). The baselines include LoRA~\citep{hu2022lora}, DoRA~\citep{liu2024dora}, BOFT~\citep{boft}, VeRA~\citep{kopiczko2023vera}, and SVFT~\citep{lingam2024svft}. Full experimental details are provided in Appendix~\ref{appendix:hyperparam}.

\paragraph{Mathematical Reasoning}

In Table~\ref{tab:math_reasoning}, we provide results on mathematical question answering, comparing our method against baseline PEFT methods across three different base models ranging from 2B to 8B parameters. Our experiments include two configurations of PiCa: a high-rank setting with fewer trainable parameters than SVFT$^R$, and a low-rank configuration with fewer trainable parameters than rank 1 LoRA. As shown in Table~\ref{tab:math_reasoning}, our high-rank PiCa consistently achieves superior performance while using the fewest trainable parameters across all models and datasets. In the low-rank setting, PiCa achieves either the best or second-best performance.

\begin{table*}[ht!]
\centering
\small
\addtolength{\tabcolsep}{-4.8pt}
\caption{Performance on Mathematical Reasoning benchmarks (GSM-8K and MATH). \#Params indicates the number of trainable parameters. The {best} and {second-best} PEFT methods are highlighted in \textbf{bold} and \underline{underlined}, respectively. For Gemma-7B, we set $r=16$ to ensure the number of trainable parameters remains below that of rank-1 LoRA. For SVFT$_d^R$, we use $d=16$ for Gemma models and $d=12$ for LLaMA-3 models. In the high-rank setting, {PiCa consistently achieves the best performance across all models and datasets, while using the fewest trainable parameters}.}

\vspace{0.05cm}
\begin{tabular}{lccccccccc}
\toprule
\multirow{2}{*}{\textbf{Method}\rule{0pt}{3ex}} & \multicolumn{3}{c}{Gemma-2B} & \multicolumn{3}{c}{Gemma-7B} & \multicolumn{3}{c}{LLaMA-3-8B} \\
\cmidrule(l){2-4} \cmidrule(l){5-7} \cmidrule(l){8-10}
 & \textbf{\#Params} & \textbf{GSM-8K} & \textbf{MATH} & \textbf{\#Params} & \textbf{GSM-8K} & \textbf{MATH} & \textbf{\#Params} & \textbf{GSM-8K} & \textbf{MATH} \\
\midrule \midrule
Full-FT & 2.5B & 52.69 & 17.94 & 8.5B & {78.09} & {30.98} & 8.0B & {76.57} & {26.12} \\
\midrule
\(\text{BOFT}^{b=8}_{m=2}\) & 1.22M & 36.01 & 12.13 & 2.90M & 71.79 & \textbf{28.98} & 4.35M & 67.09 & 21.64 \\
\(\text{DoRA}_{r=1}\) & 1.19M & 35.35 & 13.04 & 3.26M & \textbf{74.37} & 26.28 & 2.55M & 68.30 & \underline{21.96} \\
\(\text{LoRA}_{r=1}\) & 0.82M & 32.97 & 13.04 & 0.82M & 72.40 & 26.28 & 1.77M & 68.84 & 20.94 \\
\(\text{VeRA}_{r=1024}\) & 0.63M & 36.77 & 14.12 & 0.43M & 71.11 & 27.04 & 0.98M & 63.76 & 20.28 \\
\textsc{SVFT}$^P$ & 0.19M & \underline{40.34} & \underline{14.38} & 0.43M & 73.50 & 27.30 & 0.48M & \underline{69.22} & 20.44 \\
\rowcolor{rowgray} PiCa$_{r=32}$ & 0.67M & \textbf{41.32} & \textbf{15.22} & 0.64M & \underline{74.30} & \underline{28.92} & 1.38M & \textbf{73.54} & \textbf{24.14} \\
\midrule\midrule
\(\text{LoRA}_{r=32}\) & 26.2M & 43.06 & 15.50 & 68.8M & 76.57 & 29.34 & 56.6M & 75.89 & \underline{24.74} \\
\(\text{DoRA}_{r=16}\) & 13.5M & 44.27 & \underline{16.18} & 35.5M & 74.52 & 29.84 & 29.1M & 75.66 & 24.72 \\
\textsc{SVFT}$^R_{\text{d}}$ & 6.35M & \underline{50.03} & 15.56 & 19.8M & \underline{76.81} & \underline{29.98} & 13.1M & \underline{75.90} & 24.22 \\
\rowcolor{rowgray} PiCa$_{r=256}$ & 5.37M & \textbf{52.77} & \textbf{16.36} & 10.22M & \textbf{78.39} & \textbf{30.16} & 11.01M & \textbf{76.12} & \textbf{24.88} \\

\bottomrule
\end{tabular}
\label{tab:math_reasoning}
\end{table*}

\begin{table*}[h!]
\centering
\caption{Performance on Commonsense Reasoning benchmarks. \#Params refers to the number of trainable parameters. The best and second-best PEFT methods are highlighted in \textbf{bold} and \underline{underlined} text, respectively. In the high-rank setting, PiCa achieves state-of-the-art performance on 7 out of 8 datasets, using over 13$\times$ fewer parameters than LoRA and about half the parameters of SVFT.}
\vspace{0.05cm}
\resizebox{\textwidth}{!}{
\begin{tabular}{lrccccccccc}
\toprule
\textbf{Method} & \textbf{\#Params} & \textbf{BoolQ} & \textbf{PIQA} & \textbf{SIQA} & \textbf{HS} & \textbf{WG} & \textbf{ARC-e} & \textbf{ARC-c} & \textbf{OBQA} & \textbf{Avg.} \\
\midrule
\midrule
Full-FT              & 8.5B   & 72.32 & 87.32 & 76.86 & 91.07 & {81.76} & 92.46 & 82.87 & {89.00} & 84.19 \\
\cmidrule(lr){1-11}
DoRA$_{r=1}$           & 3.31M  & \underline{68.22} & \textbf{86.72} & 75.23 & \underline{91.14} & \textbf{78.13} & 91.87 & \textbf{83.19} & \textbf{86.20} & \underline{82.59} \\
VeRA$_{r=2048}$        & 1.49M  & 64.25 & 86.28 & 74.04 & 86.96 & 69.00 & \underline{92.76} & 82.33 & 82.00 & 79.70 \\
LoRA$_{r=1}$           & 0.82M  & 65.44 & 86.28 & 75.02 & 89.91 & 75.92 & 91.79 & 81.91 & 85.40 & 81.46 \\
SVFT$_P$             & 0.51M  & 67.92 & \underline{86.45} & \underline{75.47} & 86.92 & 74.03 & 91.80 & 81.21 & 83.00 & 80.85 \\
\cellcolor{rowgray}PiCa$_{r=16}$ & \cellcolor{rowgray}0.64M  & \cellcolor{rowgray}\textbf{70.95} & \cellcolor{rowgray}86.29 & \cellcolor{rowgray}\textbf{76.00} & \cellcolor{rowgray}\textbf{91.42} & \cellcolor{rowgray}\underline{76.32} & \cellcolor{rowgray}\textbf{92.89} & \cellcolor{rowgray}\textbf{83.19} & \cellcolor{rowgray}\underline{85.60} & \cellcolor{rowgray}\textbf{82.83} \\
\midrule\midrule
LoRA$_{r=32}$          & 68.8M  & 71.55 & \underline{87.95} & \underline{77.27} & 91.80 & \textbf{79.71} & \underline{92.67} & 82.16 & \underline{86.40} & \underline{83.69} \\
DoRA$_{r=16}$          & 35.5M  & 71.46 & 87.59 & 76.35 & \underline{92.11} & 78.29 & 92.00 & 80.63 & 85.60 & 83.00 \\
SVFT$_{d=8}^{B}$       & 9.80M  & \underline{71.90} & 86.96 & 76.28 & 91.55 & 78.76 & {92.80} & \underline{83.11} & 85.40 & 83.35 \\
\cellcolor{rowgray}PiCa$_{r=128}$         & \cellcolor{rowgray}5.11M  & \cellcolor{rowgray}\textbf{72.84} & \cellcolor{rowgray}\textbf{87.98} & \cellcolor{rowgray}\textbf{77.79} & \cellcolor{rowgray}\textbf{92.82} & \cellcolor{rowgray}\underline{79.40} & \cellcolor{rowgray}\textbf{93.14} & \cellcolor{rowgray}\textbf{83.62} & \cellcolor{rowgray}\textbf{88.20} & \cellcolor{rowgray}\textbf{84.47} \\
\bottomrule
\end{tabular}
}
\label{tab:common-sense-results}
\end{table*}
\begin{table*}[!h]
\centering
\small
\addtolength{\tabcolsep}{-4.2pt}
\caption{
Performance of DeBERTaV3\textsubscript{base} on the GLUE benchmark. \#Params refers to the number of trainable parameters. The best and second-best PEFT methods are highlighted in \textbf{bold} and \underline{underlined} text, respectively. While using more than 2.5$\times$ fewer parameters than $\text{SVFT}_{d=2}^{R}$, PiCa outperforms it on all datasets.
}

\vspace{0.05cm}
\begin{tabular}{lcccccccccc}\toprule
\textbf{Method} & \textbf{\#Params} & \textbf{MNLI} & \textbf{SST-2} & \textbf{MRPC} & \textbf{CoLA} & \textbf{QQP} & \textbf{QNLI} & \textbf{RTE} & \textbf{STS-B} & \textbf{Avg.}\\ \midrule \midrule
Full-FT & 183.83M & 89.90 & 95.63 & 89.46 & 69.19 & {92.40} & {94.03} & 83.75 & 91.60 & 88.25 \\ \midrule
\( \text{LoRA}_{r=8} \) & 1.33M & \textbf{90.65} & 94.95 & {89.95} & 69.82 & \textbf{93.87} & 91.99 & 85.20 & 91.60 & 88.50 \\
\( \text{LoRA}_{r=1} \) & 0.17M & 90.12 & {95.64} & 86.43 & 69.13 & 91.43 & 94.18 & 87.36 & 91.52 & 88.23 \\
\( \text{DoRA}_{r=4} \) & 0.75M & 89.92 & 95.41 & 89.10 & 69.37 & 91.53 & 94.14 & 87.00 & \underline{91.80} & 88.53 \\
\( \text{BOFT}^{b=8}_{m=2} \) & 0.75M & \underline{90.25} & \textbf{96.44} & \textbf{92.40} & \underline{72.95} & \underline{92.10} & \underline{94.23} & \underline{88.81} & \textbf{91.92} & \textbf{89.89} \\ 
\( \text{VeRA}_{r=1024} \) & 0.09M & 89.93 & 95.53 & 87.94 & 69.06 & 90.40 & 93.24 & 87.00 & 88.71 & 87.73 \\
\(\textsc{SVFT}^P\) & 0.06M & 89.69 & 95.41 & 88.77 & 70.95 & 90.16 & \textbf{94.27} & 87.24 & \underline{91.80} & 88.54  \\
 \textsc{SVFT$^R_{d=2}$} & 0.28M & 89.97 & 95.99 & 88.99 & 72.61 & 91.50 & 93.90 & 88.09 & 91.73 & 89.10 \\
\cellcolor{rowgray}PiCa$_{r=16}$ & \cellcolor{rowgray}0.11M & \cellcolor{rowgray}90.20 &\cellcolor{rowgray} \underline{96.00} & \cellcolor{rowgray}\underline{91.40} & \cellcolor{rowgray}\textbf{73.10} & \cellcolor{rowgray}{91.60} &\cellcolor{rowgray}94.20 &\cellcolor{rowgray}\textbf{89.20} &\cellcolor{rowgray}\underline{91.80} &\cellcolor{rowgray}\underline{89.69} \\
\bottomrule
\end{tabular}
\label{tab:glue_deberta_results}
\end{table*}

\begin{table*}[ht!]
\centering
\small
\addtolength{\tabcolsep}{-4.2pt}
\caption{Performance on vision benchmarks. VTAB-1K (ViT-B/16) is averaged over 19 datasets grouped into \textit{Natural, Specialized, Structured}. DreamBooth is evaluated with Stable Diffusion v2.1 using DINO (subject fidelity) and CLIP-T (text fidelity). The best and second-best results are highlighted in \textbf{bold} and \underline{underlined}, respectively.}
\vspace{0.05cm}

\begin{tabular}{lccccc @{\hspace{1.2em}} lccc}
\toprule
\multicolumn{6}{c}{\textbf{VTAB-1K (ViT-B/16)}} &
\multicolumn{4}{c}{\textbf{DreamBooth (Stable Diffusion v2.1)}} \\
\cmidrule(lr){1-6} \cmidrule(lr){7-10}
\textbf{Method} & \textbf{\#Params} & \textbf{Natural} & \textbf{Specialized} & \textbf{Structured} & \textbf{All} &
\textbf{Method} & \textbf{\#Params} & \textbf{DINO} & \textbf{CLIP-T} \\
\midrule\midrule
LoRA$_{r=8}$     & 1.32M & 0.823 & \textbf{0.851} & \textbf{0.508} & \underline{0.696} &
LoRA$_{r=16}$     & 3.37M & 0.618 & 0.305 \\
DoRA$_{r=8}$     & 1.41M & \textbf{0.827} & 0.846 & 0.505 & 0.695 &
DoRA$_{r=16}$     & 3.42M & 0.617 & \underline{0.306} \\
SVFT$^{B}_{d=8}$ & 0.93M & 0.820 & 0.844 & 0.486 & 0.684 &
SVFT$^{B}_{d=12}$ & 2.50M & \underline{0.622} & \textbf{0.307} \\
VeRA$_{r=4096}$ & 0.45M & 0.813 & 0.845 & 0.474 & 0.677 &
VeRA$_{r=13312}$ & 1.80M & 0.613 & 0.305 \\
\rowcolor{rowgray}
PiCa$_{r=64}$    & 0.44M & \underline{0.825} & \textbf{0.851} & \textbf{0.508} & \textbf{0.697} &
PiCa$_{r=128}$ & 1.72M & \textbf{0.634} & \underline{0.306} \\
\bottomrule
\end{tabular}
\label{tab:vision_results}
\end{table*}

\paragraph{Commonsense Reasoning}
In Table~\ref{tab:common-sense-results}, we evaluate commonsense reasoning performance on eight benchmark datasets using Gemma-7B, following the same experimental setup as in the Mathematical Reasoning task. We compare both high-rank and low-rank configurations of our method against PEFT baselines. In both settings, PiCa outperforms all baselines on average across the eight datasets. In the high-rank setting, our method achieves state-of-the-art performance on seven out of eight datasets while using over 13$\times$ fewer parameters than LoRA, and it consistently outperforms SVFT on all eight datasets with approximately half the number of parameters. In the low-rank setting, PiCa also achieves the best average performance, surpassing rank 1 DoRA while using more than 5$\times$ fewer parameters. Compared to SVFT$^P$, our method delivers superior performance on seven out of eight datasets, with an average improvement of nearly two percentage points. Similar trends are observed with Gemma-2B (see Appendix~\ref{sec:gemma2b}).

\paragraph{Natural Language Understanding}
Table~\ref{tab:glue_deberta_results} presents the results on the GLUE benchmark using DeBERTaV3\textsubscript{base}. Compared to LoRA with rank 8, our method achieves over one percentage point higher average performance. While using more than 2.5$\times$ fewer parameters than SVFT$_{d=2}^{R}$, our method outperforms it on all datasets. Furthermore, despite using over 7$\times$ fewer parameters than BOFT, our method achieves comparable average performance.

\paragraph{Vision Experiments}
Table~\ref{tab:vision_results} reports results on VTAB-1K and DreamBooth dataset. On the VTAB-1K dataset, PiCa achieves the best overall score while using the fewest trainable parameters. In particular, PiCa achieves competitive results compared to other baselines while using 2 to 3$\times$ fewer trainable parameters in VTAB-1K. On the DreamBooth dataset, PiCa achieves a higher DINO score while maintaining a comparable CLIP-T score, demonstrating strong personalization with fewer parameters than other baselines. These results highlight that PiCa maintains strong performance on vision tasks under substantially reduced parameter budgets.

\subsection{Further analysis}
\label{further}

\begin{wraptable}{r}{0.56\textwidth}
\centering
\caption{Ablation study on projection choice (rank = 256). 
Average scores are reported across commonsense reasoning benchmarks 
using Gemma-2B.}
\begin{tabular}{lcc}
\toprule
\textbf{Projection Method} & \textbf{\#Params} & \textbf{Avg.} \\
\midrule\midrule
Random Space & 5.37M & 63.18 \\
\rowcolor{rowgray} Column Space (Ours) & 5.37M & \textbf{67.60} \\
\bottomrule
\end{tabular}
\label{tab:projection_ablation}
\end{wraptable}

\paragraph{Ablation Study of Column Space Projection} In Table~\ref{tab:projection_ablation}, we compare the effect of using column space projection versus random space projection. We use commonsense reasoning benchmarks with Gemma-2B. The results show that column space projection improves overall accuracy by 4.42 points compared to random space projection, demonstrating the effectiveness of leveraging the spectral structure of pre-trained weights, aligned with the results in Theorem~\ref{thm:pica}.

\begin{figure}[h!]
  \centering
  
  \begin{subfigure}[t]{0.5\textwidth}
    \centering
    \includegraphics[width=\linewidth]{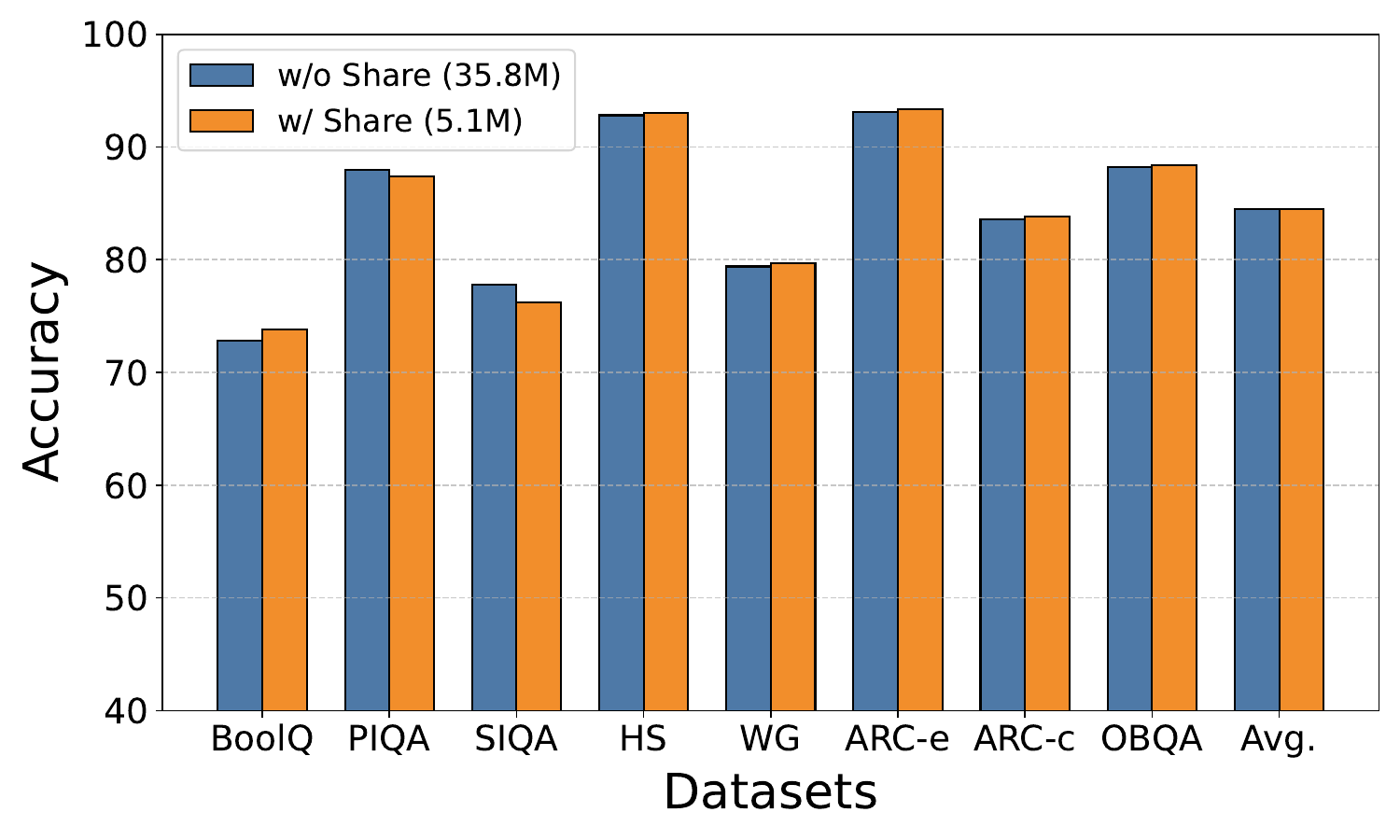}
    \caption{Accuracy across Commonsense Reasoning datasets with and without weight-sharing. weight-sharing reduces the number of trainable parameters by up to 7× without compromising performance.}
    \label{fig:sub1}
  \end{subfigure}
  \hfill
  \begin{subfigure}[t]{0.45\textwidth}
    \centering
    \includegraphics[width=\linewidth]{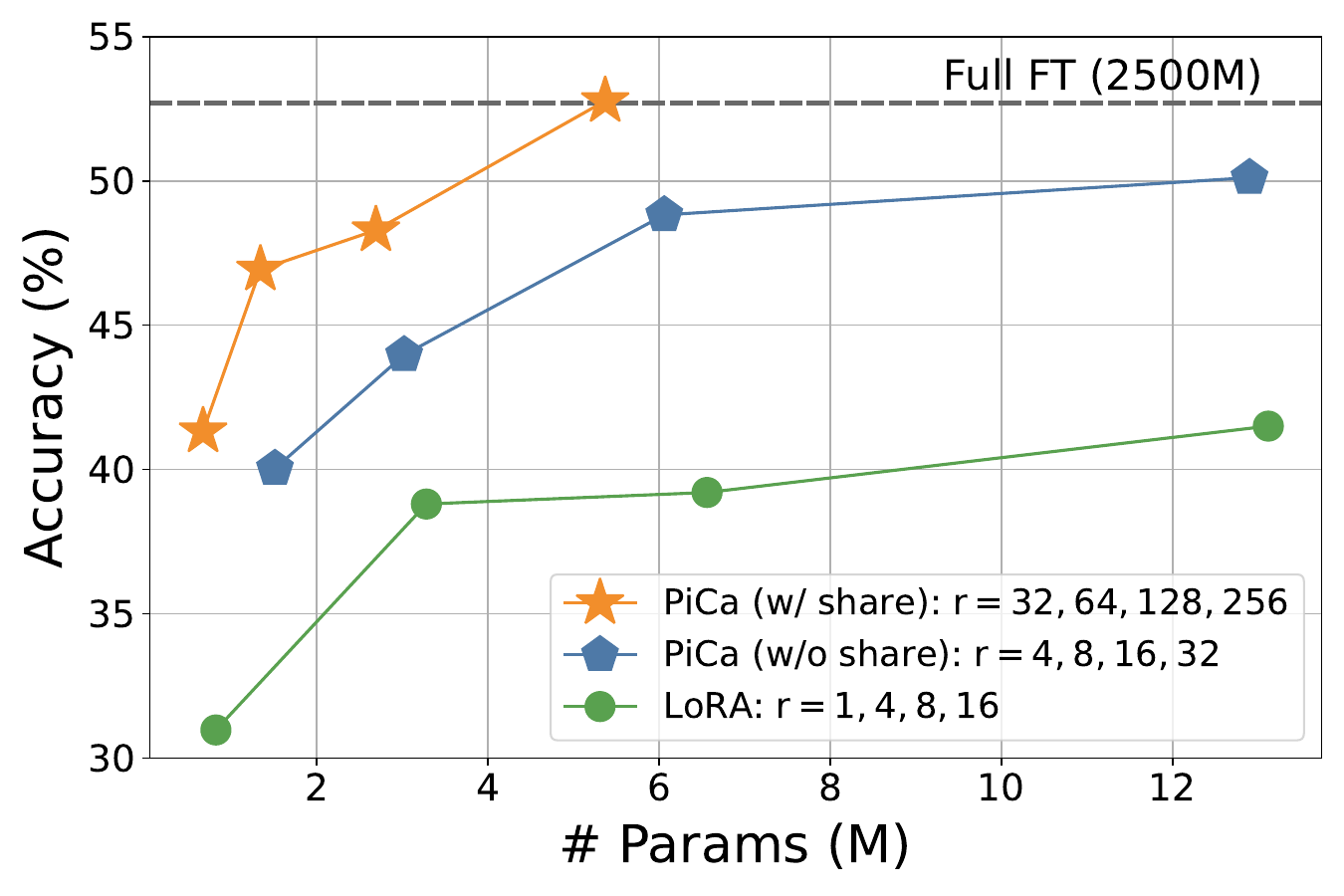}
    \caption{Accuracy on GSM-8K under varying rank settings. weight-sharing consistently yields superior performance under similar parameter budgets.}
    \label{fig:sub2}
  \end{subfigure}
  
  \caption{Ablation study of weight-sharing across different datasets and rank configurations.}
  \label{fig:combined}
\end{figure}

\paragraph{Ablation Study of Weight-sharing}
In Fig.~\ref{fig:sub1}, we analyze the impact of weight-sharing in PiCa across eight Commonsense Reasoning datasets using Gemma-7B. By comparing PiCa with its standard configuration (rank 128 with weight-sharing; 5.1M trainable parameters) against a variant without sharing (rank 16; 35.8M parameters), we find that the default PiCa consistently achieves performance comparable to its non-sharing variant while requiring about 7$\times$ fewer trainable parameters.These results indicate that weight-sharing substantially improves parameter efficiency without performance degradation.

Furthermore, we conduct an additional study on the effect of weight-sharing under varying rank settings using the GSM-8K benchmark with Gemma-2B. As shown in Fig.~\ref{fig:sub2}, PiCa consistently achieves superior performance under similar parameter budgets compared to both its no-sharing ablation and LoRA.

\section{Discussion}
While PiCa significantly reduces the number of trainable parameters required, it introduces a minor limitation during inference. Specifically, PiCa stores only a small shared matrix \( B^f \) for each functional group $f$, but requires to perform an additional SVD on the pre-trained weights \( W_0 \) at load time to recover the projection matrix \( P^{f,i}=U^{f,i} \). This presents a trade-off between storage cost and loading overhead. If the loading overhead is a concern, one can optionally store \(U^{f,i} \). Nonetheless, in scenarios where multiple task-specific adaptations are required from a single base model, PiCa offers greater scalability: a shared set of task-agnostic \(U^{f,i} \) can be pre-computed and paired with multiple sets of lightweight task-specific \( B^f \), enabling efficient adaptation across diverse tasks.

\section{Conclusion}
In this work, we introduced PiCa, a parameter-efficient fine-tuning method that integrates gradient projection onto the principal column space of pre-trained weights with a novel weight-sharing mechanism. Our theoretical analysis establishes that column space projection provides an effective inductive bias for fine-tuning, while the addition of weight-sharing offers substantial reductions in trainable parameters without compromising performance. Through extensive experiments, we demonstrated that PiCa consistently achieves competitive or superior results compared to state-of-the-art baselines across a wide spectrum of NLP tasks (Mathematical Reasoning, Commonsense Reasoning, and Natural Language Understanding) as well as challenging vision tasks (Visual Adaptation and Subject-Driven Generation). 

Taken together, our results indicate that PiCa offers a theoretically grounded and empirically validated approach to parameter-efficient adaptation of large models. We hope this work motivates further exploration of theoretically guided approaches that unify geometry-aware design with practical efficiency in fine-tuning large-scale foundation models. In future work, we aim to extend PiCa to more dynamic and practical settings such
as multi-task adaptation and continual learning, where efficient and scalable fine-tuning is critical.

\section*{Acknowledgment}
This work was supported by the National Research Foundation of Korea (NRF) grant funded by the Korea government (MSIT) (No. RS-2024-00345809, Research on AI Robustness Against Distribution Shift in Real-World Scenarios; and No. RS-2023-00222663, Center for Optimizing Hyperscale AI Models and Platforms).

% \newpage
{
\bibliography{reference}

@article{lingam2024svft,
  title={Svft: Parameter-efficient fine-tuning with singular vectors},
  author={Lingam, Vijay Chandra and Neerkaje, Atula and Vavre, Aditya and Shetty, Aneesh and Gudur, Gautham Krishna and Ghosh, Joydeep and Choi, Eunsol and Dimakis, Alex and Bojchevski, Aleksandar and Sanghavi, Sujay},
  journal={Advances in Neural Information Processing Systems},
  volume={37},
  pages={41425--41446},
  year={2024}
}

@inproceedings{mantri2025ditask,
  title={DiTASK: Multi-Task Fine-Tuning with Diffeomorphic Transformations},
  author={Mantri, Krishna Sri Ipsit and Sch{\"o}nlieb, Carola-Bibiane and Ribeiro, Bruno and Baskin, Chaim and Eliasof, Moshe},
  booktitle={Proceedings of the Computer Vision and Pattern Recognition Conference},
  pages={25218--25229},
  year={2025}
}

@article{chen2024punica,
  title={Punica: Multi-tenant lora serving},
  author={Chen, Lequn and Ye, Zihao and Wu, Yongji and Zhuo, Danyang and Ceze, Luis and Krishnamurthy, Arvind},
  journal={Proceedings of Machine Learning and Systems},
  volume={6},
  pages={1--13},
  year={2024}
}

@inproceedings{liu2024dora,
  title={Dora: Weight-decomposed low-rank adaptation},
  author={Liu, Shih-Yang and Wang, Chien-Yi and Yin, Hongxu and Molchanov, Pavlo and Wang, Yu-Chiang Frank and Cheng, Kwang-Ting and Chen, Min-Hung},
  booktitle={Forty-first International Conference on Machine Learning},
  year={2024}
}

@article{shuttleworth2024lora,
  title={Lora vs full fine-tuning: An illusion of equivalence},
  author={Shuttleworth, Reece and Andreas, Jacob and Torralba, Antonio and Sharma, Pratyusha},
  journal={arXiv preprint arXiv:2410.21228},
  year={2024}
}

@article{hu2022lora,
  title={Lora: Low-rank adaptation of large language models.},
  author={Hu, Edward J and Shen, Yelong and Wallis, Phillip and Allen-Zhu, Zeyuan and Li, Yuanzhi and Wang, Shean and Wang, Lu and Chen, Weizhu and others},
  journal={ICLR},
  volume={1},
  number={2},
  pages={3},
  year={2022}
}

@article{renduchintala2023tied,
  title={Tied-LoRA: Enhancing parameter efficiency of LoRA with weight tying},
  author={Renduchintala, Adithya and Konuk, Tugrul and Kuchaiev, Oleksii},
  journal={arXiv preprint arXiv:2311.09578},
  year={2023}
}

@inproceedings{press-wolf-2017-using,
    title = "Using the Output Embedding to Improve Language Models",
    author = "Press, Ofir  and
      Wolf, Lior",
    editor = "Lapata, Mirella  and
      Blunsom, Phil  and
      Koller, Alexander",
    booktitle = "Proceedings of the 15th Conference of the {E}uropean Chapter of the Association for Computational Linguistics: Volume 2, Short Papers",
    month = apr,
    year = "2017",
    address = "Valencia, Spain",
    publisher = "Association for Computational Linguistics",
    url = "https://aclanthology.org/E17-2025/",
    pages = "157--163"
}

@article{inan2016tying,
  title={Tying word vectors and word classifiers: A loss framework for language modeling},
  author={Inan, Hakan and Khosravi, Khashayar and Socher, Richard},
  journal={arXiv preprint arXiv:1611.01462},
  year={2016}
}

@article{meng2024pissa,
  title={Pissa: Principal singular values and singular vectors adaptation of large language models},
  author={Meng, Fanxu and Wang, Zhaohui and Zhang, Muhan},
  journal={Advances in Neural Information Processing Systems},
  volume={37},
  pages={121038--121072},
  year={2024}
}

@inproceedings{houlsby2019parameter,
  title={Parameter-efficient transfer learning for NLP},
  author={Houlsby, Neil and Giurgiu, Andrei and Jastrzebski, Stanislaw and Morrone, Bruna and De Laroussilhe, Quentin and Gesmundo, Andrea and Attariyan, Mona and Gelly, Sylvain},
  booktitle={International conference on machine learning},
  pages={2790--2799},
  year={2019},
  organization={PMLR}
}

@article{song2024sharelora,
  title={ShareLoRA: Parameter Efficient and Robust Large Language Model Fine-tuning via Shared Low-Rank Adaptation},
  author={Song, Yurun and Zhao, Junchen and Harris, Ian G and Jyothi, Sangeetha Abdu},
  journal={arXiv preprint arXiv:2406.10785},
  year={2024}
}

@misc{
shen2024sharelora,
title={ShareLo{RA}: Less Tuning, More Performance for Lo{RA} Fine-tuning of {LLM}s},
author={Zheyu Shen and Guoheng Sun and Yexiao He and Ziyao Wang and Yuning Zhang and Souvik Kundu and Eric P. Xing and Hongyi Wang and Ang Li},
year={2024},
url={https://openreview.net/forum?id=O6QZ4W6GXt}
}

@misc{
zhou2025bishare,
title={Bi-Share Lo{RA}: Enhancing the Parameter Efficiency of Lo{RA} with Intra-Layer and Inter-Layer Sharing},
author={Yuhua Zhou and Ruifeng Li and Changhai Zhou and Fei Yang and Aimin PAN},
year={2025},
url={https://openreview.net/forum?id=Thv66GmqZS}
}

@article{kopiczko2023vera,
  title={Vera: Vector-based random matrix adaptation},
  author={Kopiczko, Dawid J and Blankevoort, Tijmen and Asano, Yuki M},
  journal={arXiv preprint arXiv:2310.11454},
  year={2023}
}

@inproceedings{han2023svdiff,
  title={Svdiff: Compact parameter space for diffusion fine-tuning},
  author={Han, Ligong and Li, Yinxiao and Zhang, Han and Milanfar, Peyman and Metaxas, Dimitris and Yang, Feng},
  booktitle={Proceedings of the IEEE/CVF International Conference on Computer Vision},
  pages={7323--7334},
  year={2023}
}

@article{qiu2023controlling,
  title={Controlling text-to-image diffusion by orthogonal finetuning},
  author={Qiu, Zeju and Liu, Weiyang and Feng, Haiwen and Xue, Yuxuan and Feng, Yao and Liu, Zhen and Zhang, Dan and Weller, Adrian and Sch{\"o}lkopf, Bernhard},
  journal={Advances in Neural Information Processing Systems},
  volume={36},
  pages={79320--79362},
  year={2023}
}

@inproceedings{boft,
 title={Parameter-Efficient Orthogonal Finetuning via Butterfly Factorization},
 author={Liu, Weiyang and Qiu, Zeju and Feng, Yao and Xiu, Yuliang and Xue, Yuxuan and Yu, Longhui and Feng, Haiwen and Liu, Zhen and Heo, Juyeon and Peng, Songyou and Wen, Yandong and Black, Michael J. and Weller, Adrian and Sch{\"o}lkopf, Bernhard},
 booktitle={The Twelfth International Conference on Learning Representations},
 year={2024}
}

@InProceedings{mnli,
  author = "Williams, Adina
            and Nangia, Nikita
            and Bowman, Samuel",
  title = "A Broad-Coverage Challenge Corpus for 
           Sentence Understanding through Inference",
  booktitle = "Proceedings of the 2018 Conference of 
               the North American Chapter of the 
               Association for Computational Linguistics:
               Human Language Technologies, Volume 1 (Long
               Papers)",
  year = "2018",
  publisher = "Association for Computational Linguistics",
  pages = "1112--1122",
  location = "New Orleans, Louisiana",
  url = "http://aclweb.org/anthology/N18-1101"
}

@inproceedings{hellaswag,
    title={HellaSwag: Can a Machine Really Finish Your Sentence?},
    author={Zellers, Rowan and Holtzman, Ari and Bisk, Yonatan and Farhadi, Ali and Choi, Yejin},
    booktitle ={Proceedings of the 57th Annual Meeting of the Association for Computational Linguistics},
    year={2019}
}

@article{llama3,
  title = {Introducing Meta Llama 3: The most capable openly available LLM to date},
  author = {Meta AI},
  year = {2024},
  month = {April},
  url = {https://ai.meta.com/blog/meta-llama-3/}
}

@article{gemma,
  title={Gemma: Open models based on gemini research and technology},
  author={Team, Gemma and Mesnard, Thomas and Hardin, Cassidy and Dadashi, Robert and Bhupatiraju, Surya and Pathak, Shreya and Sifre, Laurent and Rivi{\`e}re, Morgane and Kale, Mihir Sanjay and Love, Juliette and others},
  journal={arXiv preprint arXiv:2403.08295},
  year={2024}
}

@inproceedings{glue,
    title = "{GLUE}: A Multi-Task Benchmark and Analysis Platform for Natural Language Understanding",
    author = "Wang, Alex  and
      Singh, Amanpreet  and
      Michael, Julian  and
      Hill, Felix  and
      Levy, Omer  and
      Bowman, Samuel",
    booktitle = "Proceedings of the 2018 {EMNLP} Workshop {B}lackbox{NLP}: Analyzing and Interpreting Neural Networks for {NLP}",
    month = nov,
    year = "2018",
    address = "Brussels, Belgium",
    publisher = "Association for Computational Linguistics",
    url = "https://aclanthology.org/W18-5446",
    doi = "10.18653/v1/W18-5446",
    pages = "353--355",
}

@article{gsm8k,
  title={Training Verifiers to Solve Math Word Problems},
  author={Cobbe, Karl and Kosaraju, Vineet and Bavarian, Mohammad and Chen, Mark and Jun, Heewoo and Kaiser, Lukasz and Plappert, Matthias and Tworek, Jerry and Hilton, Jacob and Nakano, Reiichiro and Hesse, Christopher and Schulman, John},
  journal={arXiv preprint arXiv:2110.14168},
  year={2021}
}

@misc{yu2023metamath,
      title={MetaMath: Bootstrap Your Own Mathematical Questions for Large Language Models}, 
      author={Longhui Yu and Weisen Jiang and Han Shi and Jincheng Yu and Zhengying Liu and Yu Zhang and James T. Kwok and Zhenguo Li and Adrian Weller and Weiyang Liu},
      year={2023},
      eprint={2309.12284},
      archivePrefix={arXiv},
      primaryClass={cs.CL}
}

@misc{winogrande,
      title={WinoGrande: An Adversarial Winograd Schema Challenge at Scale}, 
      author={Keisuke Sakaguchi and Ronan Le Bras and Chandra Bhagavatula and Yejin Choi},
      year={2019},
      eprint={1907.10641},
      archivePrefix={arXiv},
      primaryClass={cs.CL}
}

@misc{debertav3,
      title={DeBERTaV3: Improving DeBERTa using ELECTRA-Style Pre-Training with Gradient-Disentangled Embedding Sharing}, 
      author={Pengcheng He and Jianfeng Gao and Weizhu Chen},
      year={2023},
      eprint={2111.09543},
      archivePrefix={arXiv},
      primaryClass={cs.CL}
}

@misc{boolq,
      title={BoolQ: Exploring the Surprising Difficulty of Natural Yes/No Questions}, 
      author={Christopher Clark and Kenton Lee and Ming-Wei Chang and Tom Kwiatkowski and Michael Collins and Kristina Toutanova},
      year={2019},
      eprint={1905.10044},
      archivePrefix={arXiv},
      primaryClass={cs.CL}
}

@misc{socialiqa,
      title={SocialIQA: Commonsense Reasoning about Social Interactions}, 
      author={Maarten Sap and Hannah Rashkin and Derek Chen and Ronan LeBras and Yejin Choi},
      year={2019},
      eprint={1904.09728},
      archivePrefix={arXiv},
      primaryClass={cs.CL}
}

@misc{openbookqa,
      title={Can a Suit of Armor Conduct Electricity? A New Dataset for Open Book Question Answering}, 
      author={Todor Mihaylov and Peter Clark and Tushar Khot and Ashish Sabharwal},
      year={2018},
      eprint={1809.02789},
      archivePrefix={arXiv},
      primaryClass={cs.CL}
}

@misc{dosovitskiy2021imageworth16x16words,
      title={An Image is Worth 16x16 Words: Transformers for Image Recognition at Scale}, 
      author={Alexey Dosovitskiy and Lucas Beyer and Alexander Kolesnikov and Dirk Weissenborn and Xiaohua Zhai and Thomas Unterthiner and Mostafa Dehghani and Matthias Minderer and Georg Heigold and Sylvain Gelly and Jakob Uszkoreit and Neil Houlsby},
      year={2021},
      eprint={2010.11929},
      archivePrefix={arXiv},
      primaryClass={cs.CV},
      url={https://arxiv.org/abs/2010.11929}, 
}

@misc{rombach2022highresolutionimagesynthesislatent,
      title={High-Resolution Image Synthesis with Latent Diffusion Models}, 
      author={Robin Rombach and Andreas Blattmann and Dominik Lorenz and Patrick Esser and Björn Ommer},
      year={2022},
      eprint={2112.10752},
      archivePrefix={arXiv},
      primaryClass={cs.CV},
      url={https://arxiv.org/abs/2112.10752}, 
}

@misc{ruiz2023dreamboothfinetuningtexttoimage,
      title={DreamBooth: Fine Tuning Text-to-Image Diffusion Models for Subject-Driven Generation}, 
      author={Nataniel Ruiz and Yuanzhen Li and Varun Jampani and Yael Pritch and Michael Rubinstein and Kfir Aberman},
      year={2023},
      eprint={2208.12242},
      archivePrefix={arXiv},
      primaryClass={cs.CV},
      url={https://arxiv.org/abs/2208.12242}, 
}

@misc{zhai2020largescalestudyrepresentationlearning,
      title={A Large-scale Study of Representation Learning with the Visual Task Adaptation Benchmark}, 
      author={Xiaohua Zhai and Joan Puigcerver and Alexander Kolesnikov and Pierre Ruyssen and Carlos Riquelme and Mario Lucic and Josip Djolonga and Andre Susano Pinto and Maxim Neumann and Alexey Dosovitskiy and Lucas Beyer and Olivier Bachem and Michael Tschannen and Marcin Michalski and Olivier Bousquet and Sylvain Gelly and Neil Houlsby},
      year={2020},
      eprint={1910.04867},
      archivePrefix={arXiv},
      primaryClass={cs.CV},
      url={https://arxiv.org/abs/1910.04867}, 
}

@article{cho2024hollowed,
  title={Hollowed net for on-device personalization of text-to-image diffusion models},
  author={Cho, Wonguk and Choi, Seokeon and Das, Debasmit and Reisser, Matthias and Kim, Taesup and Yun, Sungrack and Porikli, Fatih},
  journal={Advances in Neural Information Processing Systems},
  volume={37},
  pages={43058--43079},
  year={2024}
}

@misc{hendrycks2021measuring,
      title={Measuring Mathematical Problem Solving With the MATH Dataset}, 
      author={Dan Hendrycks and Collin Burns and Saurav Kadavath and Akul Arora and Steven Basart and Eric Tang and Dawn Song and Jacob Steinhardt},
      year={2021},
      eprint={2103.03874},
      archivePrefix={arXiv},
      primaryClass={cs.LG}
}

@article{wedin1972perturbation,
  title={Perturbation bounds in connection with singular value decomposition},
  author={Wedin, Per-{\AA}ke},
  journal={BIT Numerical Mathematics},
  volume={12},
  number={1},
  pages={99--111},
  year={1972},
  publisher={Springer}
}

@inproceedings{piqa,
  author = {Yonatan Bisk and Rowan Zellers and
            Ronan Le Bras and Jianfeng Gao
            and Yejin Choi},
  title = {PIQA: Reasoning about Physical Commonsense in
           Natural Language},
  booktitle = {Thirty-Fourth AAAI Conference on
               Artificial Intelligence},
  year = {2020},
}

@misc{arc,
      title={Think you have Solved Question Answering? Try ARC, the AI2 Reasoning Challenge}, 
      author={Peter Clark and Isaac Cowhey and Oren Etzioni and Tushar Khot and Ashish Sabharwal and Carissa Schoenick and Oyvind Tafjord},
      year={2018},
      eprint={1803.05457},
      archivePrefix={arXiv},
      primaryClass={cs.AI}
}

@article{eckart1936approximation,
  title={The approximation of one matrix by another of lower rank},
  author={Eckart, Carl and Young, Gale},
  journal={Psychometrika},
  volume={1},
  number={3},
  pages={211--218},
  year={1936},
  publisher={Springer}
}

@article{weyl1912asymptotische,
  title={Das asymptotische verteilungsgesetz der eigenwerte linearer partieller differentialgleichungen (mit einer anwendung auf die theorie der hohlraumstrahlung)},
  author={Weyl, Hermann},
  journal={Mathematische Annalen},
  volume={71},
  pages={441--479},
  year={1912},
  publisher={Springer}
}
\bibliographystyle{iclr2026_conference}
}

\newpage
\appendix
\section*{Appendix}

\newtheorem{notation}{Notation}
\section{Preliminaries}
\label{sec:prelim}
\subsection{Notation}
\begin{notation}
The following notation is used throughout this paper:
\begin{itemize}
    \item For any matrix \( A \in \mathbb{R}^{m \times n} \), let \( \sigma_i(A) \) denote its \( i \)-th largest singular value, with \( \sigma_1(A) \geq \sigma_2(A) \geq \cdots \geq \sigma_{\min(m,n)}(A) \geq 0 \).
    \item \( \|A\|_F \): Frobenius norm of matrix \( A \), defined as \( \|A\|_F = \sqrt{\sum_{i,j} A_{ij}^2} \).
    \item \( \|A\|_2 \): Spectral norm of matrix \( A \), defined as \( \|A\|_2 = \sigma_1(A) \).
    \item \( A_{ij} \): Entry at the \( i \)-th row and \( j \)-th column of matrix \( A \).
    \item \( I_k \): Identity matrix of size \( k \times k \).
    \item \( \text{diag}(a_1, \dots, a_n) \): Diagonal matrix with entries \( a_1, \dots, a_n \).   
    \item \( \sin\Theta(U_r,U_r^*) \): denotes the principal angles between the subspaces 
          $\mathrm{range}(U_r)$ and $\mathrm{range}(U_r^*)$.
\end{itemize}
\end{notation}

\subsection{Preliminary Results}

\begin{lemma}[Weyl’s Inequality {\citep{weyl1912asymptotische}}]
For $A,B \in \mathbb{R}^{m \times n}$, and all $i$,
\[
|\sigma_i(A + B) - \sigma_i(A)| \leq \|B\|_2.
\]
\end{lemma}

\begin{lemma}[Invariance of Frobenius Norm]
If $A \in \mathbb{R}^{m \times n}$, and $U,V$ are orthogonal matrices, then
\[
\|UAV^T\|_F = \|A\|_F.
\]
\end{lemma}

\begin{lemma}[Orthogonal projection is non-expansive in Frobenius norm]
\label{lem:proj-nonexpansive-F}
Let $U_r\in\mathbb{R}^{m\times r}$ have orthonormal columns and let $\Pi_{U_r}=U_rU_r^\top$ be the orthogonal projector onto $\mathrm{range}(U_r)$. Then, for all $X\in\mathbb{R}^{m\times n}$,
\[
\|\Pi_{U_r}X\|_F \;\le\; \|X\|_F
\]

\end{lemma}

\section{Proof of Theorems}
\setcounter{theorem}{0}
\begin{theorem}[Approximation error of projection onto $U_r$]
Let \( W_0 = U \Sigma V^\top \in \mathbb{R}^{m \times n} \) be the Singular Value Decomposition (SVD) of \( W_0 \).  
Suppose the fine-tuned matrix \( W^* \in \mathbb{R}^{m \times n} \) has the form
\[
W^* = (U P) \Sigma^* (V Q)^\top,
\]
where:
\begin{itemize}
    \item \(U^*=UP \) and \(V^* =VQ \) are the left and right singular vectors of \( W^* \), respectively,
    \item \( \Sigma^* = \text{diag}(\sigma_1(W^*), \dots, \sigma_{\min(m,n)}(W^*)) \),
    \item \( P = I_m + E^P \), \( Q = I_n + E^Q \), with \( |E^P_{ij}| < \epsilon \), \( |E^Q_{ij}| < \epsilon \).

\end{itemize}
Let \( \Delta W = W^* - W_0 \), and let \( U_r \in \mathbb{R}^{m \times r} \) be the top-\( r \) left singular vectors of \( W_0 \).  
Then, the approximation error incurred by projecting \( \Delta W \) onto the subspace spanned by \( U_r \) satisfies
\[
\left\| \Delta W - U_r U_r^\top \Delta W \right\|_F^2 \leq \sum_{i=r+1}^{\min(m,n)} \sigma_i^2(\Delta W) + \mathcal{O}(\epsilon).
\]
\end{theorem}

\begin{proof}
\label{sec:proof}
We derive the inequality through a series of steps, decomposing the perturbation, analyzing the projection error, and bounding the terms using spectral and entrywise techniques.

The perturbed matrix has the form
\[
W^* = U (I_m + E^P) \Sigma^* (I_n + E^Q)^\top V^\top.
\]
Subtracting $W_0 = U\Sigma V^\top$ gives
\[
\Delta W = U \left[(I_m + E^P)\Sigma^*(I_n + E^Q)^\top - \Sigma\right] V^\top.
\]
For notational clarity, define
\[
H = (I_m + E^P)\Sigma^*(I_n + E^Q)^\top - \Sigma,
\]
so that $\Delta W = UHV^\top$.

Let us expand $H$ explicitly. Multiplying out terms yields
\[
(I_m + E^P)\Sigma^*(I_n + E^Q)^\top = \Sigma^* + E^P \Sigma^* + \Sigma^*(E^Q)^\top + E^P \Sigma^*(E^Q)^\top.
\]
Thus
\[
H = D + E_1 + E_2 + E_3,
\]
where
\[
D = \Sigma^* - \Sigma, \quad E_1 = E^P \Sigma^*, \quad E_2 = \Sigma^* (E^Q)^\top, \quad E_3 = E^P \Sigma^* (E^Q)^\top.
\]
The diagonal matrix $D$ captures the shifts in singular values: $D_{ii} = \sigma_i(W^*) - \sigma_i(W_0)$.

The error of projecting $\Delta W$ onto $U_r$ is
\[
\|\Delta W - U_r U_r^\top \Delta W\|_F^2.
\]
Since $\Delta W = UHV^\top$ and $U_r^\top U = [I_r\;0]$, we can write
\[
U_r U_r^\top \Delta W = U \begin{bmatrix} I_r & 0 \\ 0 & 0 \end{bmatrix} H V^\top.
\]
Subtracting gives
\[
\Delta W - U_r U_r^\top \Delta W = U(H - P_rH)V^\top,
\]
where $P_r = \begin{bmatrix} I_r & 0 \\ 0 & 0 \end{bmatrix}$.
By invariance of the Frobenius norm,
\[
\|\Delta W - U_r U_r^\top \Delta W\|_F^2 = \|H - P_r H\|_F^2 = \sum_{i=r+1}^m \sum_{j=1}^n H_{ij}^2.
\]

For $i>r$, each entry has the form
\[
H_{ij} = D_{ij} + E_{1,ij} + E_{2,ij} + E_{3,ij}.
\]

For diagonal terms ($j=i$), we have
\[
H_{ii} = \sigma_i(W^*) - \sigma_i(W_0) + E^P_{ii}\sigma_i(W^*) + \sigma_i(W^*)E^Q_{ii} + \sum_{k}E^P_{ik}\sigma_k(W^*)E^Q_{ik}.
\]

Using $|E^P_{ij}|,|E^Q_{ij}| < \epsilon$, we can bound each component:
\[
|E_{1,ii}|\le \epsilon\sigma_i(W^*),\quad
|E_{2,ii}|\le \epsilon\sigma_i(W^*),\quad
|E_{3,ii}|\le \epsilon^2 \min(m,n)\,\sigma_{\max}(W^*).
\]

For off-diagonal terms ($j\neq i$), we have
\[
H_{ij} = E^P_{ij}\sigma_j(W^*) + \sigma_i(W^*)E^Q_{ji} + \sum_k E^P_{ik}\sigma_k(W^*)E^Q_{jk},
\]

leading to analogous bounds
\[
|E_{1,ij}|\le \epsilon\sigma_j(W^*),\quad
|E_{2,ij}|\le \epsilon\sigma_i(W^*),\quad
|E_{3,ij}|\le \epsilon^2 \min(m,n)\,\sigma_{\max}(W^*).
\]

We now square and sum these contributions.  
For diagonals,
\[
H_{ii}^2 = (\sigma_i(W^*)-\sigma_i(W_0))^2 + 2(\sigma_i(W^*)-\sigma_i(W_0))(E_{1,ii}+E_{2,ii}+E_{3,ii}) + (E_{1,ii}+E_{2,ii}+E_{3,ii})^2.
\]
Cross term is bounded using Cauchy–Schwarz, and third quadratic term is bounded by $3(E_{1,ii}^2+E_{2,ii}^2+E_{3,ii}^2)$.  
Therefore,

\[
\sum_{i=r+1}^{\min(m,n)} H_{ii}^2 \leq \sum_{i=r+1}^{\min(m,n)} (\sigma_i(W^*) - \sigma_i(W_0))^2 + \epsilon C_{1}+ \epsilon^2 C_{2}.
\]
where
\[
C_{1} =\sum_{i=r+1}^{\min(m,n)}2|\sigma_i(W^*)-\sigma_i(W_0)|(2\sigma_i(W^*)+\epsilon\min(m,n)\sigma_{max}(W^*))
\]
\[
C_{2} = \sum_{i=r+1}^{\min(m,n)}3(2\sigma_i^2(W^*)+\epsilon^2\min(m^2,n^2) \sigma_{max}^2(W^*)\
\]
Similar expansions apply for off-diagonal terms, where only $E_1,E_2,E_3$ contribute.
For off-diagonal terms:
\[
\sum_{i=r+1}^m \sum_{\substack{j=1 \\ j \neq i}}^n H_{ij}^2 = \sum_{i=r+1}^m \sum_{\substack{j=1 \\ j \neq i}}^n(E_{1,ij} + E_{2,ij} + E_{3,ij})^2 \leq \sum_{i=r+1}^m \sum_{\substack{j=1 \\ j \neq i}}^n3 (E_{1,ij}^2 + E_{2,ij}^2 + E_{3,ij}^2) \leq \epsilon^2C_{3}.
\]
where
\[
C_{3} = \sum_{i=r+1}^m \sum_{\substack{j=1 \\ j \neq i}}^n3(\sigma_j^2(W^*) + \sigma_i^2(W^*) + \epsilon^2\min(m^2,n^2) \sigma_{max}^2(W^*))
\]

Collecting everything, the sum takes the form
\[
\sum_{i=r+1}^m \sum_{j=1}^n H_{ij}^2 \leq \sum_{i=r+1}^{\min(m,n)} (\sigma_i(W^*)-\sigma_i(W_0))^2 + \epsilon C_1 + \epsilon^2(C_2+C_3).
\]

Recall the decomposition
\[
H \;=\; D + E_1 + E_2 + E_3, 
\qquad
\Delta W \;=\; UHV^\top,
\]
so that by orthogonal invariance of singular values 
\[
\sigma_i(\Delta W) \;=\; \sigma_i(H)\quad\text{for all }i.
\]

Since $UP$ and $VQ$ are the singular-vector matrices of $W^*$, the factors $P,Q$ are orthogonal. Hence
\[
D \;=\; \Sigma^*-\Sigma \quad\Rightarrow\quad
\sigma_i(D) \;=\; \bigl|\sigma_i(W^*)-\sigma_i(W_0)\bigr| \quad(\forall i).
\]

Let $E_{\mathrm{tot}}:=E_1+E_2+E_3$. By Weyl’s inequality applied to $H=D+E_{\mathrm{tot}}$,
\[
\bigl|\sigma_i(H)-\sigma_i(D)\bigr|
\;=\;
\bigl|\sigma_i(\Delta W)-\bigl|\sigma_i(W^*)-\sigma_i(W_0)\bigr|\bigr|
\;\le\; \|E_{\mathrm{tot}}\|_2.
\]
We now bound $\|E_{\mathrm{tot}}\|_2$ piecewise. Using submultiplicativity and 
$\|E^P\|_2\le \|E^P\|_F\le \sqrt{mn}\,\epsilon$ (and similarly for $E^Q$), we get
\[
\|E_1\|_2=\|E^P\Sigma^*\|_2 \le \|E^P\|_2\,\|\Sigma^*\|_2 \le \sqrt{mn}\,\epsilon\,\sigma_{\max}(W^*),
\]
\[
\|E_2\|_2=\|\Sigma^*(E^Q)^\top\|_2 \le \|\Sigma^*\|_2\,\|E^Q\|_2 \le \sqrt{mn}\,\epsilon\,\sigma_{\max}(W^*),
\]
\[
\|E_3\|_2=\|E^P\Sigma^*(E^Q)^\top\|_2 \le \|E^P\|_2\,\|\Sigma^*\|_2\,\|E^Q\|_2 
\le mn\,\epsilon^2\,\sigma_{\max}(W^*).
\]
Therefore
\[
\|E_{\mathrm{tot}}\|_2 \;\le\; 2\sqrt{mn}\,\epsilon\,\sigma_{\max}(W^*)
\;+\; mn\,\epsilon^2\,\sigma_{\max}(W^*).
\]
 
Define
\[
\delta_i \;:=\; \sigma_i(\Delta W) \;-\; \bigl|\sigma_i(W^*)-\sigma_i(W_0)\bigr|,
\qquad
|\delta_i|\;\le\;\|E_{\mathrm{tot}}\|_2.
\]
Then
\[
\bigl|\sigma_i(W^*)-\sigma_i(W_0)\bigr|
\;=\; \sigma_i(\Delta W)-\delta_i,
\]
and squaring gives
\[
\bigl(\sigma_i(W^*)-\sigma_i(W_0)\bigr)^2
\;=\; \bigl(\sigma_i(\Delta W)-\delta_i\bigr)^2
\;\le\; \sigma_i^2(\Delta W)\;+\;2\,\sigma_i(\Delta W)\,\|E_{\mathrm{tot}}\|_2\;+\;\|E_{\mathrm{tot}}\|_2^2.
\]

Let $\ell:=\min(m,n)$. Summing for $i=r+1,\dots,\ell$,
\[
\sum_{i=r+1}^{\ell}\bigl(\sigma_i(W^*)-\sigma_i(W_0)\bigr)^2
\;\le\;
\sum_{i=r+1}^{\ell}\sigma_i^2(\Delta W)
\;+\;2\,\|E_{\mathrm{tot}}\|_2 \sum_{i=r+1}^{\ell}\sigma_i(\Delta W)
\;+\;(\ell-r)\,\|E_{\mathrm{tot}}\|_2^2.
\]
With the bound on $\|E_{\mathrm{tot}}\|_2$ just obtained, this can be written as
\[
\sum_{i=r+1}^{\ell}\bigl(\sigma_i(W^*)-\sigma_i(W_0)\bigr)^2
\;\le\;
\sum_{i=r+1}^{\ell}\sigma_i^2(\Delta W)
\;+\;\epsilon\,C_4 \;+\;\epsilon^2\,C_5,
\]
where
\[
C_4 \;=\; 2\Bigl(2\sqrt{mn}\,\sigma_{\max}(W^*)
+ mn\,\epsilon\,\sigma_{\max}(W^*)\Bigr)\sum_{i=r+1}^{\ell}\sigma_i(\Delta W),
\]
\[
C_5 \;=\; (\ell-r)\Bigl(2\sqrt{mn}\,\sigma_{\max}(W^*)
+ mn\,\epsilon\,\sigma_{\max}(W^*)\Bigr)^2.
\]

Finally, recalling the earlier analysis, we finally combine the bounds to obtain
\begin{align*}
\|\Delta W - U_r U_r^\top \Delta W\|_F^2 \leq& \sum_{i=r+1}^{\min(m,n)} \sigma_i^2(\Delta W) + \epsilon C_1 + \epsilon^2 C_2 + \epsilon^2 C_3  +  \epsilon C_4+ \epsilon^2C_5\\ =& \sum_{i=r+1}^{\min(m,n)} \sigma_i^2(\Delta W) + \epsilon C    
\end{align*}
where
\[
C= (C_1 + \epsilon C_2 + \epsilon C_3 + C_4 + \epsilon C_5)
\]

\end{proof}

\setcounter{theorem}{1}
\begin{theorem}[Sequential projection approximates accumulated projection]
Let $\ell:\mathbb{R}^{m\times n}\to\mathbb{R}$ be $L$-smooth with $\|\nabla \ell(W)\|_F\le G$.  
Define the unprojected gradient descent path
\[
Z_{t+1}=Z_t-\eta\nabla \ell(Z_t).
\]
Let the \emph{accumulated-projection} iterate be
\[
W_T
= W_0 - \eta\,\Pi_{U_r}\!\Bigl(\sum_{t=0}^{T-1}\nabla \ell(Z_t)\Bigr),
\]
and the \emph{sequential-projection} iterates
\[
P_{t+1}=P_t-\eta\,\Pi_{U_r}\nabla \ell(P_t), 
\qquad P_0=W_0,
\]
where $\Pi_{U_r}=U_r U_r^\top$ is the fixed rank-$r$ projector.  

Then, for any $T$, the difference satisfies
\[
\|W_T-P_T\|_F
\;\le\;\frac{\eta^2}{2}\,L G\,T(T-1)
+ O((\eta L T)^3).
\]
\end{theorem}

\begin{proof}
We now prove that the sequentially projected iterates closely approximate the delayed projection iterate when both use the same fixed projector $\Pi_{U_r}=U_rU_r^\top$.  
Throughout we work with the Frobenius norm, and recall from Lemma~\ref{lem:proj-nonexpansive-F} that $\Pi_{U_r}$ is non-expansive in $\|\cdot\|_F$.

The delayed projection iterate is defined by
\[
W_T^{\mathrm{delayed}}
= W_0 - \eta\,\Pi_{U_r}\!\Bigl(\sum_{t=0}^{T-1}\nabla \ell(Z_t)\Bigr),
\qquad
Z_{t+1}=Z_t-\eta\nabla \ell(Z_t).
\]
The sequentially projected iterates follow
\[
P_{t+1} = P_t - \eta\,\Pi_{U_r}\nabla \ell(P_t),
\qquad
P_0=W_0.
\]
Subtracting the two update rules yields
\[
P_T-W_T^{\mathrm{delayed}}
= -\,\eta \sum_{t=0}^{T-1}\Pi_{U_r}\bigl(\nabla \ell(P_t)-\nabla \ell(Z_t)\bigr).
\]

Taking Frobenius norms and using $\|\Pi_{U_r}\|_{F\to F}\le 1$,
\[
\|P_T-W_T^{\mathrm{delayed}}\|_F
\;\le\;\eta\sum_{t=0}^{T-1}\|\nabla \ell(P_t)-\nabla \ell(Z_t)\|_F.
\]
By Definition~\ref{def:matrix-Lsmooth}, $\ell$ is $L$-smooth w.r.t.\ $\|\cdot\|_F$, so the gradient is $L$-Lipschitz:
\[
\|\nabla \ell(P_t)-\nabla \ell(Z_t)\|_F \;\le\; L\|P_t-Z_t\|_F.
\]
Denoting $D_t=\|P_t-Z_t\|_F$, we obtain
\[
\|P_T-W_T^{\mathrm{delayed}}\|_F
\;\le\;\eta L\sum_{t=0}^{T-1}D_t.
\]

To bound $D_t$, expand one step of the deviation:
\[
\begin{aligned}
D_{t+1}
&= \|P_{t+1}-Z_{t+1}\|_F \\
&= \|P_t-\eta\Pi_{U_r}\nabla \ell(P_t)\;-\;(Z_t-\eta\nabla \ell(Z_t))\|_F \\
&= \|P_t-Z_t-\eta(\Pi_{U_r}\nabla \ell(P_t)-\nabla \ell(Z_t))\|_F.
\end{aligned}
\]
Applying the triangle inequality and splitting terms,
\[
\begin{aligned}
D_{t+1}
&\le D_t
   + \eta\,\|\Pi_{U_r}(\nabla \ell(P_t)-\nabla \ell(Z_t))\|_F
   + \eta\,\|(I-\Pi_{U_r})\nabla \ell(Z_t)\|_F.
\end{aligned}
\]

For the first term, by non-expansiveness of $\Pi_{U_r}$ and $L$-smoothness,
\[
\|\Pi_{U_r}(\nabla \ell(P_t)-\nabla \ell(Z_t))\|_F
\;\le\;\|\nabla \ell(P_t)-\nabla \ell(Z_t)\|_F
\;\le\; L D_t.
\]
For the second term, since $\|\nabla \ell(Z_t)\|_F\le G$ by assumption,
\[
\|(I-\Pi_{U_r})\nabla \ell(Z_t)\|_F
\;\le\;\|\nabla \ell(Z_t)\|_F
\;\le\; G.
\]
Hence the recurrence is
\[
D_{t+1}\;\le\;(1+\eta L)D_t+\eta G.
\]

With $D_0=0$, a standard unrolling argument gives
\[
D_t \;\le\;\frac{G}{L}\big((1+\eta L)^t-1\big)
\;\le\;\frac{G}{L}\big(e^{\eta L t}-1\big).
\]
Plugging back into Step 2,
\[
\|P_T-W_T^{\mathrm{delayed}}\|_F
\;\le\;\eta L\sum_{t=0}^{T-1}D_t
\;\le\;\eta G\sum_{t=0}^{T-1}(e^{\eta L t}-1).
\]
For small $\eta L T$, we use the second-order Taylor expansion of the exponential:
\[
e^{x}-1 = x + \tfrac{x^2}{2} + O(x^3)\quad\text{as }x\to 0.
\]
Applying this with $x=\eta L t$ yields
\[
e^{\eta L t}-1 \;=\; \eta L t + \tfrac{1}{2}(\eta L t)^2 + O\!\big((\eta L t)^3\big),
\]
and hence
\[
\eta L\sum_{t=0}^{T-1}D_t
\;\le\;\eta G\sum_{t=0}^{T-1}\!\bigl(e^{\eta L t}-1\bigr)
\;=\;\frac{\eta^2 L G}{2}\,T(T-1) \;+\; O\!\big((\eta L T)^3\big).
\]
Combining all estimates, we conclude
\[
\|W_T-P_T\|_F
\;\le\;\frac{\eta^2 L G}{2}\,T(T-1)\;+\;O\!\big((\eta L T)^3\big),
\]
which shows that the sequential projection scheme faithfully tracks the delayed projection up to higher-order error in the learning rate and horizon.

\end{proof}

\section{Implementation Details and Additional Experiments}
\label{appendix:hyperparam}

To ensure a direct and unbiased comparison with existing baseline methods, we adopted the same experimental setup as outlined in SVFT~\citep{lingam2024svft} for NLP tasks. For consistency, all baseline results in NLP tasks were also sourced from \citep{lingam2024svft}, enabling a fair evaluation of our method’s performance. For vision tasks, we follow \cite{dosovitskiy2021imageworth16x16words} and \cite{cho2024hollowed}. 

For Mathematical Reasoning tasks, we reproduced Full Fine-Tuning (Full FT) experiments for Gemma-7B and LLaMA-3-8B using smaller learning rates. While previous results were reported in \citep{lingam2024svft}, we found that with sufficiently low learning rates (1e-6/5e-6),  Full FT achieves higher accuracy than PEFT methods. We suspect SVFT used higher learning rates (1e-5/5e-5) for Full FT. The comparison is summarized in Table~\ref{tab:appendix_fullft_repro}.

\begin{table}[h]
\centering
\caption{Full Fine-Tuning (Full FT) accuracy on GSM8K and MATH.}
\label{tab:appendix_fullft_repro}
\begin{tabular}{ccccc}
\toprule
\textbf{Setting} 
& \multicolumn{2}{c}{\textbf{Gemma-7B}} 
& \multicolumn{2}{c}{\textbf{LLaMA-3-8B}} \\
\cmidrule(lr){2-3} \cmidrule(lr){4-5}
& GSM8K & MATH & GSM8K & MATH \\
\midrule\midrule
Reported in SVFT & 74.67 & 25.70 & 64.13 & 16.24 \\
\rowcolor{rowgray}
Our Reproduction (Low LR) & \textbf{78.09} & \textbf{30.98} & \textbf{76.57} & \textbf{26.12} \\
\bottomrule
\end{tabular}
\end{table}

\subsection{Implementation Details}
\label{app:language}

\paragraph{Mathematical Reasoning}
Table~\ref{tab:math_hparams} presents the hyperparameter configurations employed for these experiments. For the Gemma model family, PiCa is applied to the $Q, K, V, U, D$ matrices, while for the LLaMA-3-8B model, the $Q, K, V, U, D, O, G$ matrices are targeted. The experimental codebase and evaluation procedures are adapted from \url{https://github.com/VijayLingam95/SVFT.git}, and the fine-tuning dataset are sourced from \url{https://huggingface.co/datasets/meta-math/MetaMathQA-40K}.
\begin{table*}[h]
\centering
\caption{Hyperparameter setup used for fine-tuning on MetaMathQA-40K.}
\begin{tabular}{lcccccc}
\toprule
\textbf{Hyperparameter} & \multicolumn{2}{c}{\textbf{Gemma-2B}} & \multicolumn{2}{c}{\textbf{Gemma-7B}} & \multicolumn{2}{c}{\textbf{LLaMA-3-8B}} \\
\midrule\midrule
Optimizer & \multicolumn{6}{c}{AdamW} \\
Warmup Ratio & \multicolumn{6}{c}{0.1} \\
LR Schedule & \multicolumn{6}{c}{Cosine} \\
Max Seq. Len. & \multicolumn{6}{c}{512} \\
\# Epochs & \multicolumn{6}{c}{2} \\
Batch Size & \multicolumn{6}{c}{64} \\
Rank & 32 & 256 & 16 & 256 & 32 & 256 \\
Learning Rate & 1E-03 & 9E-04 & 1E-04 & 5E-05 & 2E-04 & 2E-04 \\

\bottomrule
\end{tabular}
\label{tab:math_hparams}
\end{table*}

\paragraph{Commonsense Reasoning} We follow the setting outlined in prior work \citep{lingam2024svft}, fine-tuning on 15K examples. The hyperparameter configurations for these experiments are detailed in ~\autoref{tab:commonsense_hparams}. We utilize the same set of matrices as in the Mathematical Reasoning tasks. The codebase, including training and evaluation data, is sourced from \url{https://github.com/VijayLingam95/SVFT.git}.
\begin{table*}[h]
\centering
\caption{Hyperparameter setup used for fine-tuning on commonsense-15K.}

\begin{tabular}{lcccc}
\toprule
\textbf{Hyperparameter} & \multicolumn{2}{c}{\textbf{Gemma-2B}} & \multicolumn{2}{c}{\textbf{Gemma-7B}} \\
\midrule\midrule
Optimizer & \multicolumn{4}{c}{AdamW} \\
Warmup Steps & \multicolumn{4}{c}{100} \\
LR Schedule & \multicolumn{4}{c}{Linear} \\
Max Seq. Len. & \multicolumn{4}{c}{512} \\
\# Epochs & \multicolumn{4}{c}{3} \\
Batch Size & \multicolumn{4}{c}{64} \\
Rank & 32 & 256 & 16 & 128 \\
Learning Rate & 1E-03 & 9E-04 & 3E-04 & 8E-05 \\
\bottomrule
\end{tabular}
\label{tab:commonsense_hparams}
\end{table*}

\paragraph{Natural Language Understanding} We fine-tune DeBERTaV3\textsubscript{base}~\citep{debertav3}, applying PiCa to all linear layers within each transformer block. We constrain hyperparameter optimization to moderate adjustments of the learning rate and the number of training epochs. For rigorous comparison, we employ identical model sequence lengths to those reported by \citep{lingam2024svft, boft}. The precise hyperparameter settings utilized in these experiments are specified in Table~\ref{tab:glue_deberta_hparams}.
\begin{table*}[h]
\centering
\caption{Hyperparameter setup used for DeBERTaV3\textsubscript{base} on the GLUE benchmark.}
\resizebox{0.99\textwidth}{!}{
\begin{tabular}{llcccccccc}\toprule
\textbf{Method} & \textbf{Dataset} & \textbf{MNLI} & \textbf{SST-2} & \textbf{MRPC} & \textbf{CoLA} & \textbf{QNLI} & \textbf{QQP} & \textbf{RTE} & \textbf{STS-B}\\ \midrule\midrule
& Optimizer & \multicolumn{ 8}{c}{AdamW} \\
& Warmup Ratio & \multicolumn{ 8}{c}{0.1} \\
& LR Schedule & \multicolumn{ 8}{c}{Linear} \\ 
& Batch Size & \multicolumn{ 8}{c}{32} \\ 
& Max Seq. Len. & 256 & 128 & 320 & 64 & 512 & 320 & 320 & 128 \\ 
\midrule
\multirow{ 2}{*}{PiCa$_{r=16}$} & Learning Rate & 3E-04 & 1E-03 & 2E-03 & 8E-4 & 3E-04 & 1E-04 & 1E-03 & 3E-03 \\
& \# Epochs & 5 & 7 & 35 & 50 & 5 & 15 & 40 & 15 \\ 
\bottomrule
\end{tabular}
}
\label{tab:glue_deberta_hparams}
\end{table*}

\paragraph{Vision Experiments}
For vision adaptation tasks, we fine-tune ViT-B/16~\citep{dosovitskiy2021imageworth16x16words} by updating all linear layers within each transformer block, using a learning rate of 0.004 for PiCa and LoRA, 0.005 for DoRA, and 0.05 for VeRA and SVFT.
For all methods, the classifier learning rate is fixed at $0.005$. Fine-tuning is conducted for 10 epochs, and the checkpoint from the best validation epoch is used for testing. The same hyperparameter configurations are applied across all 19 datasets of VTAB-1K~\citep{zhai2020largescalestudyrepresentationlearning}. For subject-driven generation tasks, we follow training and evaluation protocols of previous works \citep{lingam2024svft, cho2024hollowed}. We use a learning rate of 0.0001 for LoRA and DoRA, 0.0005 for PiCa, 0.001 for SVFT, and 0.005 for VeRA. Other settings remain the same with \cite{cho2024hollowed}.

\subsection{Evidence from large-scale models.} 
\label{app:evidence_llama}
While Fig.~\ref{fig:fig3} provides visual evidence of subspace alignment in moderate-scale settings, here we empirically validate the assumptions underlying Theorem~\ref{thm:pica} on a larger model. 
Specifically, we analyze LLaMA3-8B fine-tuned on Commonsense Reasoning benchmarks. 

For each pair of pre-trained and fine-tuned weight matrices, we computed the cosine similarity between their singular vectors and defined \emph{Diagonal Similarity} as the average of the diagonal entries of the similarity matrix, aggregated across layers of each module (query, key, and value). 
The consistently high Diagonal Similarity values reported in Table~\ref{tab:llama8b} demonstrate that the leading singular subspaces remain well aligned after fine-tuning, thus supporting the subspace stability assumption of Theorem~\ref{thm:pica}. 

We also extend the analysis of Fig.~\ref{fig:fig3} by reporting the averaged entries of $E^P$ and $E^Q$ across layers. 
As shown in Table~\ref{tab:llama8b}, these values are tightly concentrated around zero, empirically confirming that the additional $\mathcal{O}(\epsilon)$ term in Theorem~\ref{thm:pica} is negligible in practice.

\begin{table}[h]
\centering
\caption{Empirical validation of Theorem~\ref{thm:pica} assumptions on LLaMA3-8B fine-tuned for Commonsense Reasoning. 
Diagonal Similarity measures alignment of singular vectors between pre-trained and fine-tuned weights. 
The averaged values of $E^P_{ij}$ and $E^Q_{ij}$ are tightly concentrated near zero, confirming that the $\mathcal{O}(\epsilon)$ term is negligible.}
\label{tab:llama8b}
\begin{tabular}{lccc}
\toprule
\textbf{Layer} & \textbf{Diagonal Similarity} & \textbf{$E^P_{ij}$} & \textbf{$E^Q_{ij}$} \\
\midrule\midrule
Query & $0.927 \pm 0.047$ & $-2.44e{-}4 \pm 4.27e{-}6$ & $-2.44e{-}4 \pm 4.25e{-}6$ \\
Key   & $0.998 \pm 0.003$ & $-9.66e{-}4 \pm 3.76e{-}5$ & $-9.66e{-}4 \pm 3.76e{-}5$ \\
Value & $0.972 \pm 0.011$ & $-9.69e{-}4 \pm 2.76e{-}5$ & $-9.66e{-}4 \pm 2.76e{-}5$ \\
\bottomrule
\end{tabular}
\end{table}

\subsection{Commonsense Reasoning with Gemma-2B}
\label{sec:gemma2b}
We evaluate PiCa on commonsense reasoning tasks with Gemma-2B. The results are presented in Table~\ref{tab:common-sense-results-2b}. PiCa achieves the highest average performance across both high- and low-rank settings, outperforming the second-best method by approximately 2--3 percentage points.
\begin{table*}[ht]
\centering
\caption{Performance on Commonsense Reasoning benchmarks using Gemma-2B. \#Params refers to the number of trainable parameters. The best and second-best PEFT methods are highlighted in \textbf{bold} and \underline{underlined} text, respectively. PiCa achieves state-of-the-art average performance across both high- and low-rank settings, outperforming the second-best method by up to 3 percentage points.}
\vspace{0.05cm}
\resizebox{\textwidth}{!}{
\begin{tabular}{lccccccccccc}
\toprule
\textbf{Method} & \textbf{\#Params} & \textbf{BoolQ} & \textbf{PIQA} & \textbf{SIQA} & \textbf{HS} & \textbf{WG} & \textbf{ARC-e} & \textbf{ARC-c} & \textbf{OBQA} & \textbf{Avg.} \\
\midrule\midrule
Full-FT              & 2.5B     & 63.57 & 74.10 & 65.86 & 70.00 & 61.95 & 75.36 & 59.72 & 69.00 & 67.45 \\
\cmidrule(lr){1-11}
\(\text{BOFT}^{b=8}_{m=2}\)      & 1.22M  & 59.23 & 63.65 & 47.90 & 29.93 & 50.35 & 59.04 & 42.66 & 41.00 & 49.22 \\
VeRA$_{r=2048}$        & 0.66M  & 62.11 & 64.31 & 49.18 & 32.00 & 50.74 & 58.08 & 42.83 & 42.60 & 50.23 \\
LoRA$_{r=1}$           & 0.82M  & \underline{62.20} & 69.31 & 56.24 & 32.47 & \textbf{51.53} & \underline{69.52} & 48.80 & 56.40 & \underline{55.81} \\
DoRA$_{r=1}$           & 1.19M  & 62.17 & 68.77 & 55.93 & \underline{32.95} & \underline{51.22} & 68.81 & 48.72 & 55.60 & 55.52 \\
SVFT$_P$             & 0.19M  & \textbf{62.26} & \underline{70.18} & \underline{56.70} & 32.47 & 47.04 & 69.31 & \underline{50.08} & \underline{58.40} & \underline{55.81} \\
\rowcolor{rowgray} PiCa$_{r=32}$ & 0.67M  & 62.11 & \textbf{71.76} & \textbf{60.13} & \textbf{36.49} & 50.59 & \textbf{73.74} & \textbf{52.56} & \textbf{63.20} & \textbf{58.82} \\
\midrule\midrule
LoRA$_{r=32}$          & 26.2M  & {63.11} & 73.44 & 63.20 & 47.79 & 52.95 & 74.78 & 57.16 & 67.00 & 62.43 \\
DoRA$_{r=16}$          & 13.5M  & 62.87 & \underline{73.93} & \textbf{65.34} & 53.16 & 55.51 & \underline{76.43} & \textbf{59.55} & \textbf{68.40} & 64.40 \\
SVFT$_B^{d=16}$      & 6.35M  & \underline{63.42} & 73.72 & 63.86 & \underline{71.21} & \underline{59.58} & 73.69 & 54.77 & 66.60 & \underline{65.86}\\
\rowcolor{rowgray} PiCa $_{r=256}$         & 5.37M  & \textbf{63.91} & \textbf{75.57} & \underline{64.38} & \textbf{71.75} & \textbf{60.62} & \textbf{77.44} & \underline{58.70} & \textbf{68.40} & \textbf{67.60} \\
\bottomrule
\end{tabular}
}
\label{tab:common-sense-results-2b}
\end{table*}

\subsection{Resource and Efficiency Analysis}
\label{app:resource_analysis}

\begin{table*}[t]
\centering
\caption{Trainable parameters and training memory consumption for different parameter-efficient fine-tuning methods on Gemma-7B.}
\vspace{0.1cm}
\small
\begin{tabular}{lccccccc}
\toprule
\textbf{Method} &
LoRA$_{r=32}$ &
DoRA$_{r=16}$ &
BOFT$^{b=8}_{m=2}$ &
VeRA$_{r=1024}$ &
SVFT$^{P}$ &
SVFT$^{R}_{d=16}$ &
{PiCa$_{r=128}$} \\
\midrule\midrule
{\# Params} &
68.8M &
35.5M &
2.90M &
0.43M &
0.43M &
19.8M &
{5.11M} \\
\rowcolor{rowgray}{Memory} &
52{,}764 &
55{,}334 &
65{,}423 &
55{,}980 &
70{,}176 &
71{,}421 &
{52{,}614} \\
\bottomrule
\end{tabular}
\label{memory}
\end{table*}

We present a comparative analysis of training memory usage between PiCa and the state-of-the-art baseline SVFT.
Training memory is measured by peak GPU memory usage during fine-tuning.
As shown in Table~\ref{memory}, although both SVFT$^P$ and SVFT$^R_{d=16}$ substantially reduce the number of trainable parameters, they require up to 36\% higher GPU memory consumption compared to PiCa$_{r=128}$.

This overhead arises from SVFT’s factorization of weight updates as $\Delta W = U M V^\top$, where $U \in \mathbb{R}^{m \times m}$ and $V \in \mathbb{R}^{n \times n}$ denote the singular vectors of the pre-trained weight matrix $W_0 \in \mathbb{R}^{m \times n}$.
Although $U$ and $V$ are not trainable, they must be retained throughout fine-tuning, leading to significant memory overhead.
These results demonstrate that PiCa offers a more memory-efficient alternative, particularly in resource-constrained environments where memory I/O constitutes a critical bottleneck.

\subsection{Comparison under weight-sharing}

We compare PiCa with PEFT baselines under the same weight-sharing scheme. 
For SVFT, we report both Random and Banded variants with sharing. 
As shown in Table~\ref{tab:appendix_weight_sharing}, PiCa achieves the best performance while using fewer parameters.

\begin{table}[h]
\centering
\caption{Results with weight-sharing on GSM-8K and MATH using Gemma-2B.}
\begin{tabular}{lccc}
\toprule
\textbf{Method} & \textbf{\#Params} & \textbf{GSM-8K} & \textbf{MATH} \\
\midrule\midrule
LoRA (r=32, $B$ shared) & 14.83M & 44.28 & 15.08 \\
DoRA (r=16, $B$ shared) & 7.79M & 41.02 & 14.92 \\
SVFT\_R w/ sharing ($d=280$) & 5.48M & 50.49 & 15.86 \\
SVFT\_B w/ sharing ($d=280$) & 5.48M & 50.34 & 16.18 \\
\rowcolor{rowgray} PiCa (r=256) & 5.37M & \textbf{52.77} & \textbf{16.36} \\
\bottomrule
\end{tabular}
\label{tab:appendix_weight_sharing}
\end{table}

\section{LLM Usage}
We used large language models only for minor tasks such as spell-checking, grammar correction, and formatting.

\section{Reproducibility statement}
We have made extensive efforts to ensure the reproducibility of our work. All models, datasets, training protocols, and hyperparameters required to reproduce our experimental results are described in detail in Section~\ref{sec:exp} and Appendix~\ref{appendix:hyperparam}.

\end{document}